\documentclass[10pt,twocolumn,letterpaper]{article}

\usepackage[pagenumbers]{cvpr} 

\usepackage{graphicx}
\usepackage{amsmath}
\usepackage{amssymb} 
\usepackage{booktabs}
\usepackage{multirow}
\usepackage{enumitem}
\usepackage[accsupp]{axessibility}

\usepackage{pifont}
\newcommand{\cmark}{\ding{51}}%
\newcommand{\xmark}{\ding{55}}%

\usepackage{color, colortbl}
\definecolor{LightCyan}{rgb}{0.88,1,1}

\usepackage[table,xcdraw]{xcolor}

\newcommand{\method}{LMR}

\usepackage[pagebackref,breaklinks,colorlinks]{hyperref}

\usepackage[capitalize]{cleveref}
\crefname{section}{Sec.}{Secs.}
\Crefname{section}{Section}{Sections}
\Crefname{table}{Table}{Tables}
\crefname{table}{Tab.}{Tabs.}

\def\confName{CVPR}
\def\confYear{2023}

\begin{document}

\title{
Use Your Head: Improving Long-Tail Video Recognition \\
}
\author{Toby Perrett \hspace{20pt} Saptarshi Sinha \hspace{20pt}  Tilo Burghardt \hspace{20pt}  Majid Mirmehdi \hspace{20pt}  Dima Damen\\
{\tt\small <first>.<last>@bristol.ac.uk} \hspace{20pt}  University of Bristol, UK
}
\maketitle

\vspace{-2mm}
\begin{abstract}
This paper presents an investigation into long-tail video recognition. We demonstrate that, unlike naturally-collected video datasets and existing long-tail image benchmarks, current video benchmarks fall short on multiple long-tailed properties. Most critically, they lack few-shot classes in their tails. In response, we propose new video benchmarks that better assess long-tail recognition, by sampling subsets from two datasets: SSv2 and VideoLT.

We then propose a method, Long-Tail Mixed Reconstruction~(\method), which reduces overfitting to instances from few-shot classes by reconstructing them as weighted combinations of samples from head classes. \method\ then employs label mixing to learn robust decision boundaries. It achieves state-of-the-art average class accuracy on EPIC-KITCHENS and the proposed SSv2-LT and {VideoLT-LT}.
Benchmarks and code at: \url{tobyperrett.github.io/lmr}

\end{abstract}
\vspace{-4mm}

\section{Introduction}
\label{sec:intro}

Advances in deep learning have been driven by increasing quantities of data to train larger and more sophisticated models. Landmark recognition datasets such as ImageNet~\cite{Deng2009} and Kinetics \cite{Carreira}, amongst others, have fulfilled this need for data by first defining a taxonomy, and then scraping or crowd-sourcing until a sufficient number of examples are obtained for each class. They typically aim for balanced, or nearly balanced, class distributions.  However, in practice, collecting enough examples for every object or action, including rare ones, remains challenging. Naturally occurring data is known to come from long-tail distributions, where it is often not possible to obtain a sufficient number of samples from classes in the tail.

\begin{figure}
\centering
\includegraphics[width=0.99\columnwidth,trim={0 20 0 15}]{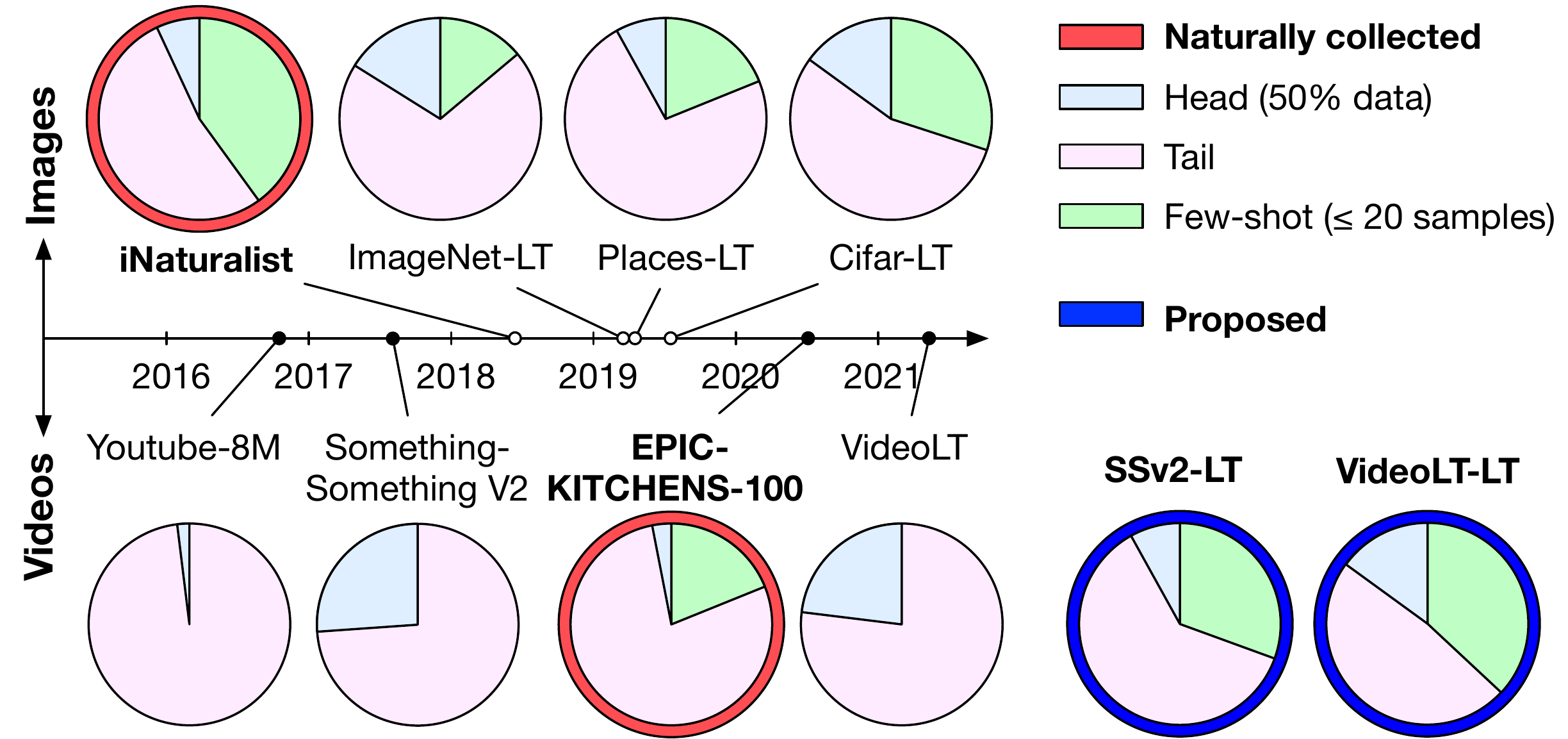}
\caption{Long-tail image recognition datasets (top) \cite{Liu2019,Cao2019a} aimed to curate similar distributions to the naturally-collected iNaturalist~\cite{Horn2018}.
For video datasets (bottom), the naturally-collected EPIC-KITCHENS-100 \cite{Damen2021} demonstrates a similar distribution of head/tail/few-shot classes to long-tail image datasets.
In comparison, curated video datasets do not include {\emph{any}} few-shot classes~\cite{Abu-El-Haija2016,Goyal2017,Zhang2021}. We propose two versions of existing datasets which do -- SSv2-LT and VideoLT-LT.
 Head/tail/few-shot definitions in \cref{fig:def}. Numeric comparison in \cref{tab:datasets}.
}
\vspace*{-12pt}
\label{fig:1}
\end{figure}

In order to encourage methods to train effectively on long-tail data, image-recognition benchmarks include multiple naturally-collected\footnote{We use the term `naturally' to focus on the data collection. It does not imply footage of nature. We hope this footnote prevents any confusion.}~\cite{Horn2018} as well as curated long-tail datasets \cite{Liu2019,Cui2019,Buda2018,Cao2019a,Wang2017a}. 
In contrast, long-tail video recognition has been a less explored field.
In~\cref{fig:1}, we compare image and video benchmarks, showcasing that none of the curated video datasets to date contain any few-shot classes~\cite{Abu-El-Haija2016,Zhang2021,Goyal2017}. 
This is a critical oversight, as seminal research has highlighted that long-tail methods must \emph{``learn accurate few-shot models for classes in the tail of the class distribution''}~\cite{Wang2017a} and \emph{``deal with imbalanced classification, few-shot learning''}~\cite{Liu2019}.
In this paper, we follow the approach from~\cite{Liu2019} and re-sample videos to introduce long-tail versions of two video datasets.

We evaluate current long-tail recognition methods on our re-sampled long-tail video datasets and the naturally-collected EPIC-KITCHENS-100 dataset~\cite{Damen2021}.
Unsurprisingly, when confronted with few-shot classes, 
current methods perform poorly due to a lack of sample diversity in the few-shot classes.
We thus propose a new method that focuses on improving the performance on few-shot classes. Long-Tail Mixed Reconstruction (\method)
reconstructs few-shot samples as weighted combinations of head samples within the batch. 
A residual connection, weighted by the class size, is used to combine instances with their reconstructions. 
We use pairwise label mixing on these reconstructed samples to help learn robust class decision boundaries.
Our key contributions are as follows:
\begin{itemize}[leftmargin=5mm,itemsep=-1.5ex,partopsep=1ex,parsep=2ex]
    \item We compare image and video long-tail datasets, by providing a consistent definition of properties for long-tail class distributions. 
    \item We curate new long-tail video benchmarks~(-LT) which better test long-tail recognition performance.
    \item We propose a method, \method, which increases the diversity in few-shot classes. It achieves highest average class accuracy across 3 benchmarks: naturally-collected EPIC-KITCHENS-100 and the two proposed curated benchmarks SSv2-LT and VideoLT-LT.
\end{itemize}

\cref{sec:def} reviews works which investigate long-tail characteristics, leading to the introduction of a set of properties and the comparison of existing long-tail benchmarks. 
\cref{sec:new_benchmarks} introduces new benchmarks and demonstrates experimentally the value of these long-tail properties.
\cref{sec:related} summarises prior long-tail and few-shot video recognition approaches. 
\cref{sec:method} introduces \method, our method for long-tail video recognition. Comparative analysis is given in \cref{sec:experiments}. Finally, ablations on \method\ are performed in \cref{sec:ablation_method}.

\begin{table*}[ht]
\resizebox{0.99\textwidth}{!}{
\footnotesize
\begin{tabular}{@{}lllrrrrrrrcl} \toprule
                                         &                                   &                                                                  & \multicolumn{1}{l}{}         & \multicolumn{3}{c}{Proposed Properties}                                                 & \multicolumn{2}{c}{Class size}                            & \multicolumn{1}{c}{Num}      & Balanced                       &                                                \\ \cline{5-7}
                                         & Source                            & Dataset                                                          & Year                         & \multicolumn{1}{c}{{H}\%}  & \multicolumn{1}{c}{{F}\%}  & \multicolumn{1}{c}{{I}}       & \multicolumn{1}{c}{Max}       & \multicolumn{1}{c}{Min}   & \multicolumn{1}{c}{classes}  & test                           & Content                                        \\ \midrule
                                         & \cellcolor[HTML]{FFD3D3}Natural   & \cellcolor[HTML]{FFD3D3}iNaturalist \cite{Horn2018}              & \cellcolor[HTML]{FFD3D3}2018 & \cellcolor[HTML]{FFD3D3}7  & \cellcolor[HTML]{FFD3D3}40 & \cellcolor[HTML]{FFD3D3}500   & \cellcolor[HTML]{FFD3D3}1000  & \cellcolor[HTML]{FFD3D3}2 & \cellcolor[HTML]{FFD3D3}8142 & \cellcolor[HTML]{FFD3D3}\cmark & \cellcolor[HTML]{FFD3D3}Photos of species      \\
                                         & Resampled                         & ImageNet-LT \cite{Liu2019}                                       & 2019                         & 16                         & 14                         & 256                           & 1280                          & 5                         & 1000                         & \cmark                         & Image recognition                              \\
                                         & Resampled                         & Places-LT \cite{Liu2019}                                         & 2019                         & 8                          & 19                         & 996                           & 4980                          & 5                         & 365                          & \cmark                         & Photos of scenes                               \\
\multirow{-4}{*}{\rotatebox{90}{Images}} & Resampled                         & Cifar-LT-100 \cite{Cui2019}                                      & 2019                         & 15                         & 30                         & 100                           & 500                           & 5                         & 100                          & \cmark                         & Image recognition                              \\ \hline
                                         & \cellcolor[HTML]{FFD3D3}Natural   & \cellcolor[HTML]{FFD3D3}EPIC-KITCHENS-100 Verbs \cite{Damen2021} & \cellcolor[HTML]{FFD3D3}2020 & \cellcolor[HTML]{FFD3D3}3  & \cellcolor[HTML]{FFD3D3}19 & \cellcolor[HTML]{FFD3D3}14848 & \cellcolor[HTML]{FFD3D3}14848 & \cellcolor[HTML]{FFD3D3}1 & \cellcolor[HTML]{FFD3D3}97   & \cellcolor[HTML]{FFD3D3}\xmark & \cellcolor[HTML]{FFD3D3}Egocentric actions     \\
                                         & Collected                         & Youtube-8M \cite{Abu-El-Haija2016}                               & 2016                         & 2                          & 0                          & 6409                          & 788288                        & 123                       & 3862                         & \xmark                         & Youtube                                        \\
                                         & Collected                         & Something-Something V2 \cite{Goyal2017}                          & 2017                         & 26                         & 0                          & 79                            & 3234                          & 41                        & 174                          & \xmark                         & Temporal reasoning                             \\
                                         & Collected                         & VideoLT \cite{Zhang2021}                                          & 2021                         & 23                         & 0                          & 43                            & 1912                          & 44                        & 1004                         & \xmark                         & Youtube (fine-grained)                         \\
                                         & \cellcolor[HTML]{C1DAFF}Resampled & \cellcolor[HTML]{C1DAFF}SSv2-LT (proposed)                       & \cellcolor[HTML]{C1DAFF}2022 & \cellcolor[HTML]{C1DAFF}9  & \cellcolor[HTML]{C1DAFF}32 & \cellcolor[HTML]{C1DAFF}500   & \cellcolor[HTML]{C1DAFF}2500  & \cellcolor[HTML]{C1DAFF}5 & \cellcolor[HTML]{C1DAFF}174  & \cellcolor[HTML]{C1DAFF}\cmark & \cellcolor[HTML]{C1DAFF}Temporal reasoning     \\
\multirow{-6}{*}{\rotatebox{90}{Videos}} & \cellcolor[HTML]{C1DAFF}Resampled & \cellcolor[HTML]{C1DAFF}VideoLT-LT (proposed)                    & \cellcolor[HTML]{C1DAFF}2022 & \cellcolor[HTML]{C1DAFF}12 & \cellcolor[HTML]{C1DAFF}38 & \cellcolor[HTML]{C1DAFF}110   & \cellcolor[HTML]{C1DAFF}550   & \cellcolor[HTML]{C1DAFF}5 & \cellcolor[HTML]{C1DAFF}772  & \cellcolor[HTML]{C1DAFF}\cmark & \cellcolor[HTML]{C1DAFF}Youtube (fine-grained) \\ \bottomrule
\end{tabular}
}
\vspace{-5pt}
\caption{Comparison of datasets against long-tail properties: Head Length ({H\%}), Few-Shot Length ({F\%}) and Imbalance~({I}). Red highlighted rows contain naturally-collected datasets. The bottom two rows (blue) contain our proposed VideoLT-LT and SSv2-LT, which are curated to better match naturally-collected data than other video benchmarks.}
\label{tab:datasets}
\vspace{-5pt}
\end{table*}

\section{Properties of Long-Tail Benchmarks} \label{sec:def}

Established benchmarks for long-tail image recognition~\cite{Liu2019} have shaped the progress of long-tail methods.
These followed earlier efforts that
investigated the desired data distribution characteristics for long-tail benchmarks.
In \cite{Buda2018}, experiments were performed with class counts that decay linearly or decay with a step-function.
They noted that a larger imbalance between majority (now known as `head') and minority (i.e. `tail') classes increases difficulty and that 
a longer tail negatively affects classifier performance for both linear and step class count decays.
Interestingly, imbalance was shown to affect higher complexity tasks (\eg CIFAR) significantly more than lower complexity tasks (\eg MNIST). 
Step and exponential class count decays were also investigated in \cite{Cao2019a}, with similar conclusions.
In \cite{Cui2019}, multiple long-tail versions of CIFAR \cite{Krizhevsky2009} were curated by changing the minimum class size. Distribution characteristics were not explored numerically, but a drop in performance was reported as the number of samples per class decreased.

\begin{figure}
\centering
\includegraphics[width=0.99\columnwidth,trim={0 10 0 5}]{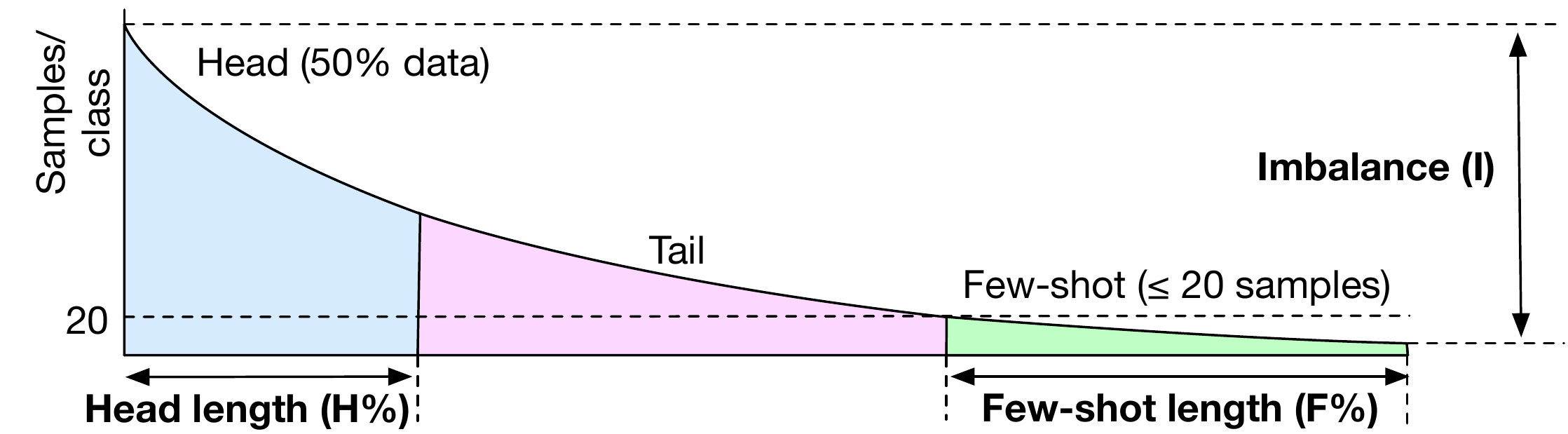}
\caption{Visualisation of long-tail distribution properties: head length ({H\%}), few-shot length ({F\%}) and imbalance ({I}). Previous works have relied solely on imbalance, or used the terms ``head'', ``mid'' and ``tail'' to describe different parts of the distribution with arbitrarily chosen sizes. In this paper, we use consistent properties to compare long-tail benchmarks across images and videos.}
\label{fig:def}
\vspace{-4mm}
\end{figure}

Despite the richness of these early findings, imbalance (\ie the ratio between the largest and smallest class sizes) has become the primary metric for characterising long-tail benchmarks. 
However, imbalance ignores other critical characteristics such as the number of few-shot classes.
To reflect this, we define three properties which together allow a more informed comparison of long-tail benchmarks. These are visualised in Fig. \ref{fig:def}:
\begin{itemize}[leftmargin=5mm,itemsep=-1.5ex,partopsep=1ex,parsep=2ex]
\item \textbf{Head Length \textbf{(H\%)}:}
The percentage of classes that formulates the majority of samples in the dataset. When classes are ranked by their size in the training set, these are the largest classes that together contribute $x$\% of the training samples. 
While different values can be used for $x$, we follow prior work that used 50\% of the data to represent head classes~\cite{Anderson,Starr2008}.
We consider the head length as the ratio of head classes to all classes.

\item \textbf{Few-Shot Length \textbf{(F\%)}:}
The percentage of few-shot classes in the dataset, where a few shot class contains $\le x$ training samples.
Prior works use values between 5 and 50 for $x$~\cite{Vinyals2016,Ravi2017,Tokmakov2019,Upervised2018,Zhu2018,Mishra,Zhang2020,Cao2020,triantafillou2019metadataset,Alayrac2022}. We follow long-tail image works and use 20 as the threshold for few-shot classes~\cite{Liu2019,Zhangb}. 

\item \textbf{Imbalance \textbf{(I)}:}
Previously used in \cite{Cui2019}, imbalance is the ratio between the size of the largest and smallest classes. 
Note that this metric alone does not provide a measure of how long-tailed a dataset is. 
\vspace{-1mm}
\end{itemize}
These three properties are distribution agnostic, \ie they can describe the properties of any benchmark whether the data is naturally-collected, or when it is sampled, no matter what distribution function is used. 
Using these three properties ({H\%, F\%, I}), we now quantitatively compare long-tail datasets across images and videos.

\subsection{Long-Tail Image Datasets}\label{sec:lt_image_datasets}

The definitive example of a naturally-collected long-tail image recognition dataset is iNaturalist 2018 \cite{Horn2018}. It is constructed from image and label contributions of plants and animals in the wild. As some species are rare, it would be very difficult to acquire more examples of these few-shot classes.
As shown in \cref{tab:datasets}, the iNaturalist image dataset has a head length of 7\% (\ie the 7\% largest classes contribute 50\% of the data), a few-shot  length of 40\% (\ie 40\% of the classes have 20 or fewer training examples) and an imbalance of 500. Thus, for methods to perform well on naturally-collected data, they must be good at learning a large number of few-shot classes.

Methods also evaluate on curated long-tail versions of large-scale datasets to avoid over-specialisation on iNaturalist. The widely used ImageNet-LT \cite{Liu2019}, Places-LT \cite{Liu2019} and CIFAR-LT \cite{Cui2019} re-sample from the original datasets and have comparable properties to the naturally-collected iNaturalist, making them suitable for evaluating methods that target long-tail recognition. As shown in \cref{tab:datasets}, these have few-shot lengths of $14\%, 19\%$ and $30\%$ respectively and head lengths of $\le 16\%$.

\subsection{Long-Tail Video Datasets}\label{sec:lt_video_datasets}

By analogy, one naturally-collected large-scale and long-tail video dataset is EPIC-KITCHENS-100 \cite{Damen2021}. Collection was unscripted recording of several days of kitchen activities. The number of samples of an action class roughly correlates to the how frequently the action occurs in daily activities. Table~\ref{tab:datasets} shows EPIC-KITCHENS-100 has a head length of 3\% and a few-shot length of 19\%. 

There have been two attempts at curating video datasets to specifically test long-tail methods, Youtube8M \cite{Abu-El-Haija2016} and VideoLT \cite{Zhang2021}. 
While these are appreciated efforts, they are far from ideal as long-tail benchmarks.
Table \ref{tab:datasets} shows neither of these contain any few-shot classes ({F\% = 0}), and VideoLT has a significantly smaller imbalance of $43$ compared to $100-996$ for long-tail image datasets.
We build on this effort to propose long-tail benchmarks that satisfy all the desired properties.

\section{Proposed Long-Tail Video Benchmarks} \label{sec:new_benchmarks}
\label{sec:proposed_datasets}
Having identified weaknesses in current benchmarks used for long-tail video recognition, we first propose to use EPIC-KITCHENS-100 as it is naturally-collected and satisfies the long-tail properties (as defined in \cref{sec:def}). We also propose to resample public video datasets, so their properties are in line with curated long-tail image datasets.

SSv2 \cite{Goyal2017} is chosen as it is widely considered to be a good test of temporal understanding and has previously been re-purposed for evaluating few-shot works~\cite{Cao2020,Zhu2020}. Similarly, VideoLT \cite{Zhang2021} targets fine-grained classes. We call these curated versions SSv2-LT and VideoLT-LT, and resample these following the recipe used in~\cite{Liu2019} for ImageNet-LT and Places-LT (sampling from a Pareto distribution with $\alpha=6$). Table \ref{tab:datasets} demonstrates these curated versions match the desired properties as visualised in \cref{fig:1}. For additional details including sampled number of instances per class, see Appendix \ref{app_datasets}. 

Before proceeding to the method, ablations are first performed at a dataset level, where different curated versions of {SSv2-LT} are compared to demonstrate the impact on long-tail properties and the effect of few-shot classes. 
Full implementation details of models and metrics will be given in \cref{sec:experiments}, but for these ablations it suffices to say that Motionformer \cite{Patrick2021} is trained with cross-entropy, reporting average class accuracy over the test set, as well as over few-shot, tail and head classes.

\subsection{Importance of Long-Tail Properties} \label{sec:fixed_imbalance}

\begin{figure}
    \centering
    \includegraphics[width=1.0\columnwidth, trim={0 27 10 8}]{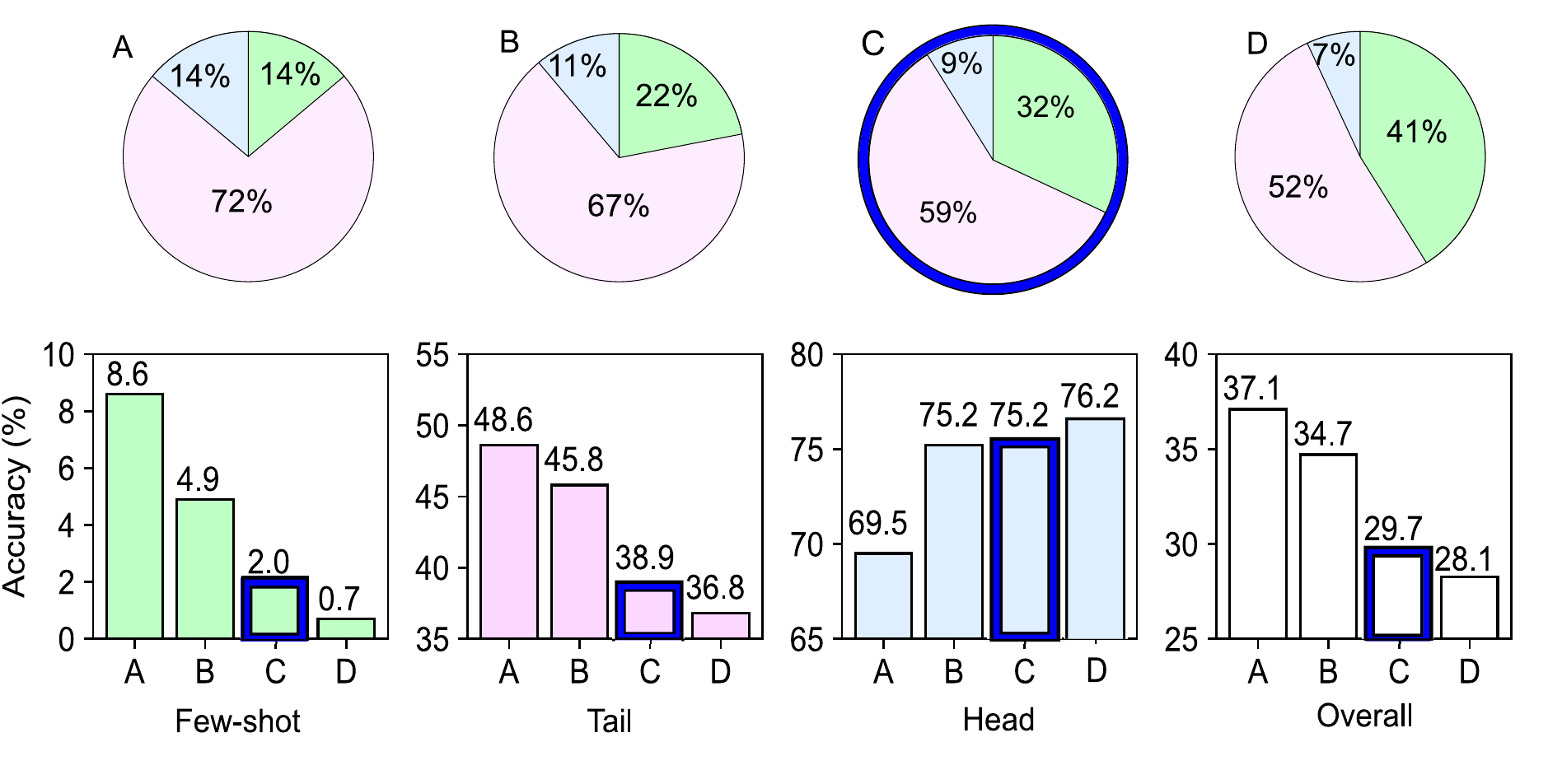}
    \caption{We compare four variants of SSv2-LT (A, B, C, D) with different {H\%} and {F\%} properties, while fixing {I} = 500, and the training dataset size = 50.4k. Top: percentage of head, tail and few-shot classes in each variant. Bottom: average class accuracy over the long-tail distribution. Variant C, highlighted in blue, is the proposed version used throughout the rest of this paper.}
    \vspace{-8pt}
    \label{fig:alpha}
\end{figure}

In \cref{sec:def}, we noted that prior works use Imbalance ({I}) to identify a dataset as being long-tailed~\cite{meta-weight-net,Cui2019}. We quantitatively showcase that imbalance alone is insufficient by constructing four variants of SSv2-LT (A, B, C, D), with a fixed training set size = 50.4k and a fixed imbalance {I} = 500. We vary the head length {H\%} and the few-shot length {F\%} as shown in \cref{fig:alpha}. Variant C (which uses an identical decay to ImageNet-LT and Places-LT \cite{Liu2019}), highlighted in blue, is the version used throughout this paper and proposed as the long-tail benchmark SSv2-LT.

As {H\%} decreases and  {F\%} increases (A $\rightarrow$ D), there are significant drops in few-shot, tail and overall accuracy (up to 9\%), whereas head performance improves. This is indicative of the distribution becoming more long-tailed. Because this behaviour occurs with fixed {I}, it can be concluded that {H\%} and {F\%} are indeed necessary for comparison of long-tail distributions.

\subsection{Effect of Few-Shot Classes}

\begin{figure}
    \centering
    \includegraphics[width=1\linewidth, trim={0 20 0 9}]{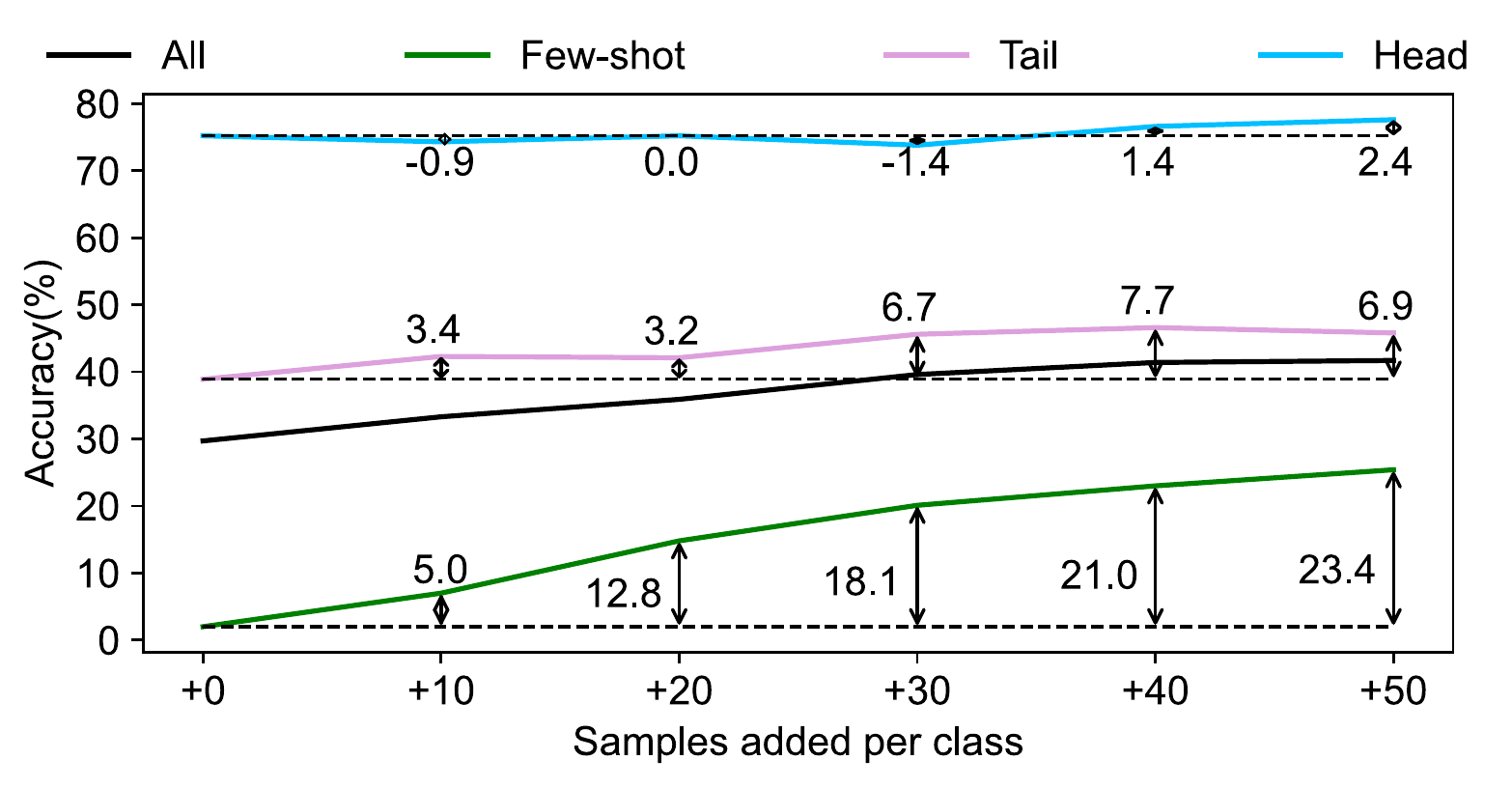}
    \caption{Effect of adding $+x$ samples per class on SSv2-LT. Average class accuracy is reported overall and for head, tail and few-shot classes. Per-case improvement reported next to arrow.} 
    \vspace{-8pt}
    \label{fig:add+constants}
\end{figure}

To showcase the importance of few shot classes, i.e. classes with $\le 20$ samples in training, we increment all classes in SSv2-LT with a fixed number of additional samples $+x$.
We evaluate the performance over few-shot/tail/head classes\footnote{We maintain the set of classes in few-Shot/tail/head for direct comparison, i.e. even if the class has $>$ 20 samples after the addition.} as we add $\{10, 20, 30, 40, 50\}$ samples per class.
\cref{fig:add+constants} shows that the accuracy on few-shot classes significantly increases when adding a small number of samples per class. The effect is smaller for tail classes and marginal for head.
The maximum improvement of few-shot classes occurs around +20 samples/class, when no few-shot classes remain in training. 
To address this challenge, it is thus important to have a sufficient number of few-shot classes in long-tail benchmarks.
Recall that naturally-collected datasets contain significant few-shot length (40\% for iNaturalist and 19\% for EPIC-KITCHENS-100).

\section{Related Methods} \label{sec:related}

Having justified our proposed benchmarks and before introducing our method, we first review long-tail and few-shot video recognition methods.

\subsection{Long-Tail Methods}
There are two main approaches to tackle long-tail recognition: re-weighting and re-balancing. 

\noindent\textbf{Re-weighting }approaches impose a higher penalty when misclassifying samples from tail classes. This can be done by directly adjusting logits \cite{Ren2020,Menon2021,Tian2020,gaussian_la,adversarial_lt} or weighting the loss by class size \cite{Cui2019,EQL_v2} or individual sample difficulty~\cite{Lin,Wangd,Park,equalized_focal,jamal_2020,meta-weight-net}. Alternative approaches include label smoothing \cite{MiSLAS} and enforcing separation between class embeddings~\cite{TSC,Samuel2021,Kasarla2022}.
Re-weighting can also be achieved by enabling more experts to specialise on tail classes and combining predictions \cite{Cui2021,Cai2021,Zhangb,Zhu2020a,trustworthy_classification,ride}.

\noindent\textbf{Re-balancing }approaches instead adjust the frequency at which examples from different classes are used in training, without adjusting the loss function. This can be achieved using a class-equalising feature bank~\cite{OLTR_TPAMI} or more commonly by equal sampling from each class~\cite{LVIS} or by instance difficulty \cite{CVPR_rebalancing,sinha_ijcv}. It has become standard practice to first train the representation with instance-balanced sampling~\cite{invariant_feature_learning} followed by class-balanced sampling~\cite{weight_balancing,Kang2020,DisAlign}.

Augmentations are known to introduce diversity into tail samples~\cite{Lia}. They combine the sample with a nearby class prototype \cite{imagine_by_reasoning}, or create feature \emph{clouds} to expand tail classes~\cite{Liu2020a}. Further augmentation approaches include combining class-specific and class-generic features~\cite{Chu}, 
using a separate classifier to identify head samples that can be adjusted and re-labeled as tail classes~\cite{Kim},
or pasting tail foreground objects onto backgrounds from head classes~\cite{CMO}. Contrastive learning has also been used to improve representations \cite{balanced_contrastive,Cuia}. For video, Framestack \cite{Zhang2021} proposes temporally mixing up samples, frame-wise, based on average-precision during training.

Our proposed method, \method, belongs to the re-balancing category. 
It is related to approaches for augmentation but differs in that it uses samples from {\emph{multiple}} other classes, weighting the reconstruction by the class count and jointly reconstructing all samples in the batch.

\subsection{Few-Shot Video Recognition}

Despite the infancy of the long-tail video recognition field, the related field of few-shot video recognition has been more widely studied \cite{Zhu2018,Dwivedi2019,Bishay2019,Zhu2020,Zhang2020,Cao2020,Zhu2021,Perrett2021,Wu2022,Wang2022,Thatipelli2022}. Instead of learning a long-tailed class distribution, few-shot methods learn to distinguish between a limited number of balanced few-shot classes (\eg 5-way 5-shot). 
Few-shot video methods rely on attention between frames of the query video and all samples in the support set of each class~\cite{Zhu2021,Wu2022,Wang2022,Perrett2021}. This requires the support set to be held in memory, which makes few-shot methods unsuitable for direct application to long-tail learning. Further, due to their design around balanced benchmarks, these methods cannot handle imbalance. 

Our method takes inspiration from few-shot works in designing an approach for long-tail video recognition. In particular, image \cite{Doersch2020} and video~\cite{Perrett2021} few-shot methods use a reconstruction technique to measure the similarity between a query and a class. 
A similar technique is used in~\cite{Patrick2021a} as input to a text captioning module. 
Each video is reconstructed from similar videos in the batch, using a cross-modal embedding space.
In contrast to these works, we apply reconstruction \emph{across classes} using multiple head samples to benefit those in the tail or those which are few shot. 
We also make use of a residual connection to maintain knowledge of the sample itself. We detail our method next.

\section{Method}\label{sec:method}

When performing class-balanced sampling, instances from the tail are oversampled. This is particularly problematic for few-shot classes, where insufficient sample diversity results in overfitting.
We propose Long-Tail Mixed Reconstruction (\method), which aims to recover this diversity by
computing a linear combination of the sample itself and weighted combinations of similar samples in the batch, weighted by the class size and followed by pairwise label mixing.
In contrast to standard augmentation techniques, reconstructions are more representative of examples likely to be seen at test time, since they make use of visually similar samples from within the training set. 

We first describe how classes are treated differently based on their count. We then proceed to describe our reconstruction and pairwise label mixing.

\subsection{Long-Tailed Class Contribution}

We consider the long-tailed class distribution of samples in the training set, and take $C_y$ as the count of the class with label $y$. 
We define a contribution function~$\mathbf{c}(y)$, per class, which we use later for reconstructing instances. 
We first calculate $\tilde C_y$ as the weight of class $y$:
\vspace{-4pt}
\begin{equation}
    \tilde C_y =  \frac{1}{\log\left(C_y d + \epsilon\right)} , \vspace{-2pt}
\end{equation}
where $d$ controls the decay, and $\epsilon$ is a constant which ensures a positive denominator. These class weights can then be used to calculate the contribution function (low for head classes, high for tail):
\vspace{-4pt}
\begin{equation}\label{eq:contribution}
    \mathbf{c}(y) =  \frac{\tilde C_y - \min(\tilde C_y)}{\max(\tilde C_y) - \min(\tilde C_y)}l ~. \vspace{-2pt}
\end{equation}

Here, $0 \le l \le 1$ is a hyperparameter controlling the contribution for the lowest class count.
Note that these class contributions are established for the classes based on the training set, and not changed during training.

\begin{figure}
\centering
\subfloat[The few-shot sample (green) is reconstructed as a weighted sum of other samples in the batch. Few-shot samples are prevented from contributing to the reconstructions through a mask (top left). \label{fig:method_batch}]{\includegraphics[scale=0.56]{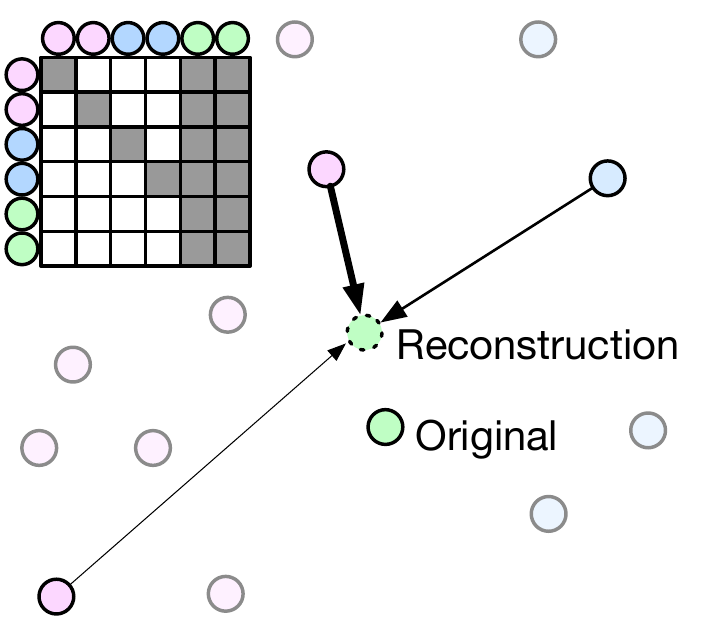}} \hspace{2mm}
\subfloat[
Reconstructions (dotted outline) increase diversity for few-shot classes, expanding class boundaries. Robust boundaries are learned by pairwise mixing of reconstructions (double outline).\label{fig:method_boundary}]{\includegraphics[scale=0.61]{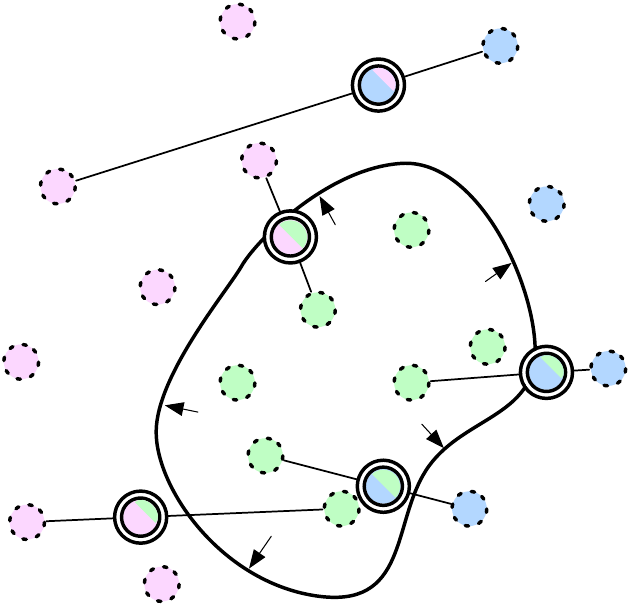}}
\vspace{-2pt}
\caption{\method\ overview: reconstruction (a) and label-mixing~(b).
}
\vspace{-5pt}
\label{fig:method}
\end{figure}

\subsection{Long-Tail Mixed Reconstruction}

\noindent \textbf{Setup.} Recognition methods combine a feature encoder $\mathbf{f}(\cdot)$ and a classifier $\mathbf{g}(\cdot)$. Data is fed to the model for training in the form of batches, where a batch $X$ contains $B$ videos $X = \{x_i: i=1...B\}$ with associated labels $Y = \{y_i: i=1...B\}$. Given the class contribution function from Eq.~\ref{eq:contribution}, we look up $\mathbf{c}(Y)$ for the samples in the batch, given their class labels.

To start, features for the batch are computed in the forward pass as ${Z = \mathbf{f}(X)}$.
We propose a mixed reconstructor $\mathbf{mr}(\cdot, \cdot)$, which acts on features $Z$ and labels $Y$, and returns a new reconstructed representation with an updated label for every video in the batch.

\noindent \textbf{Sample reconstruction.} 
We calculate cosine similarity~$\textbf{s}$ between all features within the batch, ${S_{ij} = \textbf{s} \left( Z_i, Z_j \right)}$. 
Note that here, $i$ denotes the feature to be reconstructed, and $j$ denotes the feature being used for the reconstruction. 
We then calculate an exclusion mask $E$, avoiding self-weighting, i.e. samples should not contribute to their own reconstructions, and samples from few-shot classes are also avoided as these are already oversampled. The exclusion mask $E$ is visualised in \cref{fig:method_batch}, and calculated as:
\vspace{-4pt}
\begin{equation} \label{eq:mask}
E_{ij} = 
    \begin{cases}
        0 & \text{if } (i = j)  \text{ or } (C_{y_i} \leq \omega)\\
        1 & \text{otherwise}
    \end{cases} \vspace{-2pt}
\end{equation}
where $\omega=20$ is the few-shot threshold. 
Next, we apply a softmax operation over non-masked elements per row (\ie one softmax per $i$), which calculates reconstruction weights~$W$:
\vspace{-7pt}
\begin{equation}\label{eq:softmax}
 W_{ij} = \frac{\exp(S_{ij}) E_{ij}}{ \sum_{k=1}^B \exp(S_{ik}) E_{ik}  } ~.
\end{equation}

We use a residual connection weighted by the class contribution -- the smaller the class, the more the weighted features $WZ$ contribute to the reconstruction of samples from that class. 
Specifically: 
\begin{equation} \label{eq:contribution_rep_rec}
   R = \mathbf{c}(Y)WZ + (1-\mathbf{c}(Y)) Z ~, \vspace{-2pt}
\end{equation}
where $\mathbf{c}(Y)$ (Eq.~\ref{eq:contribution}) are the contribution functions of the class labels in the batch and $R$ are the reconstructed features. 
For few shot classes, the reconstruction is mostly formed from the weighted combination of other {\emph {similar}} samples in the batch. 
Note that these reconstructions have the same class labels $Y$ as the features $Z$ they replace.

\noindent \textbf{Pairwise label mixing.} Once the reconstructions $R$ are obtained, we take a step further by performing stochastic pairwise mixing (\cref{fig:method_boundary}). 
We use a mixing mask $M$ such that:
\begin{equation} \label{eq:mixing-mask}
M_{ij} = 
    \begin{cases}
        \alpha_i & \text{if } (i = j)\\
        1-\alpha_i & \text{if } (j = \beta_i)\\
        0 & \text{otherwise}
    \end{cases} \vspace{-2pt}
\end{equation}
where $\alpha$ is a $B$-dimensional set of mixing weights, one for each sample. Following standard mixing, $\alpha_i = 1$ with probability 0.5, and randomly $0 \le \alpha_i \le 1$ otherwise.
$\beta$ is a $B$ dimensional sample selector, that selects
a different sample from the batch. ${1 \le \beta_i \le B, \beta_i \neq i \text{ and } \beta_i \in \mathbb{N}}$.

We apply the mixing mask $M$ to our reconstructions $R$ and their labels $Y$ such that
\vspace{-4pt}
\begin{equation} \label{eq:r_final}
\mathbf{mr}(Z,Y) = (M R, M Y) ~. \vspace{-2pt}
\end{equation}
We then pass these reconstructed and mixed features with the corresponding mixed labels to the classifier $\mathbf{g}$ to train.

\subsection{Training and Inference}

As customary~\cite{Kang2020}, the classifier $\mathbf{g}$, acting on the backbone $\mathbf{f}$, is first pre-trained with instance-based sampling and cross-entropy. Afterwards, $\mathbf{g}$ is reset.
\method\ is then trained with class-balanced sampling and cross-entropy on $\mathbf{g}$.
This is backpropagated through the mixed reconstructor $\mathbf{mr}$ and feature extractor $\mathbf{f}$.
At inference, $\mathbf{mr}$ is discarded, as a suitable feature extractor $\mathbf{f}$ and classifier $\mathbf{g}$ have been learned for long-tail recognition. Each test sample/video is processed independently, \ie there is no reconstruction, and labels and class counts are not used.

\section{Experiments}\label{sec:experiments}

We first perform comparative analysis on EPIC-KITCHENS-100, SSv2-LT and VideoLT-LT. 

\noindent\textbf{Metrics.}
The primary metric for long-tail video recognition is average class accuracy (Avg C/A), as it provides a fair evaluation when the test set is unbalanced.  When the test set is balanced, as in the case of SSv2-LT and VideoLT-LT, Avg C/A and overall accuracy (Acc) are identical metrics. EPIC-KITCHENS-100 has an unbalanced test set so overall accuracy is also provided for reference.
We also report average class accuracy for few-shot (marked ``few'' in tables), tail and head classes, as defined by the properties in \cref{sec:def}.

\noindent\textbf{Baselines.}
We compare against the following methods, also identified in~\cite{Zhang2021} as suitable for long-tail video recognition:
\vspace{-5pt}
\begin{itemize}[leftmargin=3mm,itemsep=-2ex,partopsep=1ex,parsep=2ex]
    \item \textbf{CE:} Standard cross entropy trained with instance-balanced sampling.
    \item \textbf{EQL:} As in \textbf{CE}, but using an Equalization Loss \cite{Tan2020}, which reduces the penalty for misclassifying a head class as a tail class. This baseline is currently used by video transformer works to address class imbalance \cite{Wu2022}.
    \item \textbf{cRT:} Classifier Retraining \cite{Kang2020}. This is now the standard practice of instance-balanced sampling, followed by a classifier reset and class-balanced sampling.
    \item \textbf{Mixup} \cite{Zhang2018a}\textbf{:}  Pairs of samples and their labels are mixed.
    \item \textbf{Framestack} \cite{Zhang2021}\textbf{:} Mixes up video frames based on a running total of class average precision.
\end{itemize}
\vspace{-5pt}

\begin{table*}
\centering
\resizebox{0.92\textwidth}{!}{
\footnotesize
\begin{tabular}{@{}l|rrrcr|rrrc|rrrc@{}}
\toprule
                            & \multicolumn{5}{c|}{\textbf{EPIC-KITCHENS-100}}                                                                                                                                                                                                    & \multicolumn{4}{c|}{\textbf{SSv2-LT}}                                                                                                                                                      & \multicolumn{4}{c}{\textbf{VideoLT-LT}}                                                                                                                                                   \\ 
Method                      & \multicolumn{1}{l}{\cellcolor[HTML]{D7FFDA}Few} & \multicolumn{1}{l}{\cellcolor[HTML]{FCD7FF}Tail} & \multicolumn{1}{l}{\cellcolor[HTML]{D7EBFF}Head} & \multicolumn{1}{l}{Avg C/A} & \multicolumn{1}{l|}{{\color[HTML]{8d8d8d} Acc}} & \multicolumn{1}{l}{\cellcolor[HTML]{D7FFDA}Few} & \multicolumn{1}{l}{\cellcolor[HTML]{FCD7FF}Tail} & \multicolumn{1}{l}{\cellcolor[HTML]{D7EBFF}Head} & \multicolumn{1}{l|}{Avg C/A = Acc} & \multicolumn{1}{l}{\cellcolor[HTML]{D7FFDA}Few} & \multicolumn{1}{l}{\cellcolor[HTML]{FCD7FF}Tail} & \multicolumn{1}{l}{\cellcolor[HTML]{D7EBFF}Head} & \multicolumn{1}{l}{Avg C/A = Acc} \\ \midrule
CE                          & \cellcolor[HTML]{D7FFDA}0.0                     & \cellcolor[HTML]{FCD7FF}12.3                     & \cellcolor[HTML]{D7EBFF}\textbf{55.2}            & 21.2                        & {\color[HTML]{8d8d8d} \textbf{63.5}}           & \cellcolor[HTML]{D7FFDA}2.0                     & \cellcolor[HTML]{FCD7FF}38.9                     & \cellcolor[HTML]{D7EBFF}\textbf{75.2}            & 29.7                              & \cellcolor[HTML]{D7FFDA}17.4                    & \cellcolor[HTML]{FCD7FF}51.1                     & \cellcolor[HTML]{D7EBFF}\textbf{75.9}            & 41.0                              \\
EQL \cite{Tan2020}          & \cellcolor[HTML]{D7FFDA}0.0                     & \cellcolor[HTML]{FCD7FF}12.4                     & \cellcolor[HTML]{D7EBFF}55.0                     & 21.1                        & {\color[HTML]{8d8d8d} 63.3}                    & \cellcolor[HTML]{D7FFDA}3.1                     & \cellcolor[HTML]{FCD7FF}39.0                     & \cellcolor[HTML]{D7EBFF}\textbf{75.2}            & 30.1                              & \cellcolor[HTML]{D7FFDA}17.4                    & \cellcolor[HTML]{FCD7FF}51.0                     & \cellcolor[HTML]{D7EBFF}75.4                     & 40.9                              \\
cRT \cite{Kang2020}         & \cellcolor[HTML]{D7FFDA}21.4                    & \cellcolor[HTML]{FCD7FF}35.0                     & \cellcolor[HTML]{D7EBFF}51.1                     & 36.9                        & {\color[HTML]{8d8d8d} 50.1}                    & \cellcolor[HTML]{D7FFDA}14.9                    & \cellcolor[HTML]{FCD7FF}45.6                     & \cellcolor[HTML]{D7EBFF}58.6                     & 36.5                              & \cellcolor[HTML]{D7FFDA}30.5                    & \cellcolor[HTML]{FCD7FF}\textbf{56.9}            & \cellcolor[HTML]{D7EBFF}64.0                     & 47.5                              \\
Mixup \cite{Zhang2018a}     & \cellcolor[HTML]{D7FFDA}25.8                    & \cellcolor[HTML]{FCD7FF}33.8                     & \cellcolor[HTML]{D7EBFF}51.7                     & 36.8                        & {\color[HTML]{8d8d8d} 51.7}                    & \cellcolor[HTML]{D7FFDA}17.4                    & \cellcolor[HTML]{FCD7FF}\textbf{46.6}            & \cellcolor[HTML]{D7EBFF}57.1                     & 37.8                              & \cellcolor[HTML]{D7FFDA}15.8                    & \cellcolor[HTML]{FCD7FF}48.9                     & \cellcolor[HTML]{D7EBFF}72.5                     & 38.9                              \\
Framestack \cite{Zhang2021} & \cellcolor[HTML]{D7FFDA}23.0                    & \cellcolor[HTML]{FCD7FF}33.6                     & \cellcolor[HTML]{D7EBFF}52.1                     & 36.5                        & {\color[HTML]{8d8d8d} 52.5}                    & \cellcolor[HTML]{D7FFDA}15.5                    & \cellcolor[HTML]{FCD7FF}46.1                     & \cellcolor[HTML]{D7EBFF}61.9                     & 37.2                              & \cellcolor[HTML]{D7FFDA}18.2                    & \cellcolor[HTML]{FCD7FF}51.8                     & \cellcolor[HTML]{D7EBFF}74.5                     & 41.5                              \\
\method                     & \cellcolor[HTML]{D7FFDA}\textbf{35.7}           & \cellcolor[HTML]{FCD7FF}\textbf{36.8}            & \cellcolor[HTML]{D7EBFF}51.1                     & \textbf{39.7}               & {\color[HTML]{8d8d8d} 51.3}                    & \cellcolor[HTML]{D7FFDA}\textbf{17.9}           & \cellcolor[HTML]{FCD7FF}46.5                     & \cellcolor[HTML]{D7EBFF}61.0                     & \textbf{38.3}                     & \cellcolor[HTML]{D7FFDA}\textbf{34.8}           & \cellcolor[HTML]{FCD7FF}56.8                     & \cellcolor[HTML]{D7EBFF}62.1                     & \textbf{48.9}                     \\ \bottomrule
\end{tabular}
} \vspace{-4pt}
\caption{
Long-tail results on EPIC-KITCHENS-100 Verbs Val set \cite{Damen2021}, SSv2-LT and VideoLT-LT. 
Note that average class accuracy (Avg C/A) is the same as overall accuracy (Acc) for balanced test sets (SSv2-LT and VideoLT-LT). EPIC-KITCHENS-100 has an unbalanced test set, so overall accuracy, which favours over-prediction of head classes, is provided for reference.
\method\ obtains the highest average class accuracy on all datasets, as well as the highest average class accuracy over few-shot classes.
}
\label{tab:results_mf}
\end{table*}

\begin{figure*}
\centering
\includegraphics[width=0.98\textwidth,trim={30 20 20 30}]{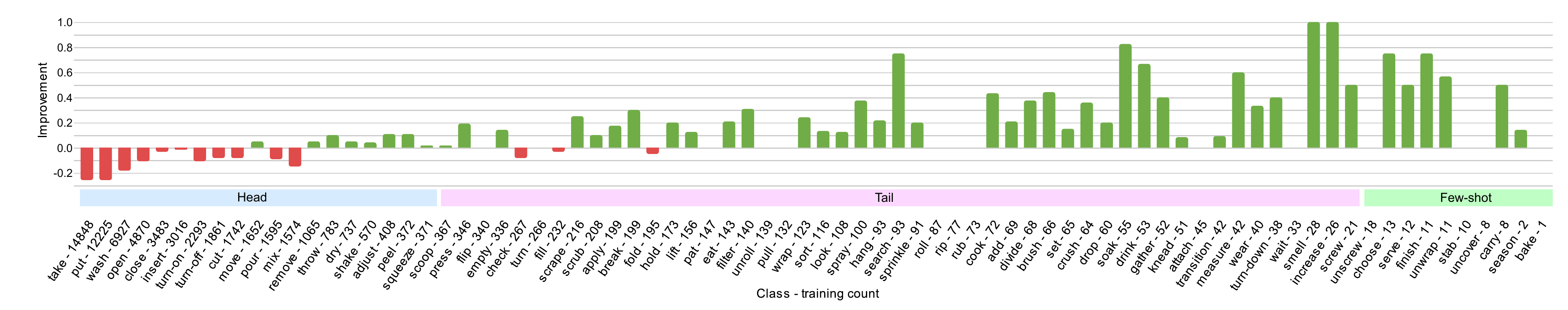}
\caption{Improvements of \method\ over CE on EPIC-KITCHENS-100. Classes are ordered by size and marked as head/tail/few-shot. \label{fig:epic_class_improvements}
} \vspace{-10pt}
\end{figure*}

\noindent\textbf{Implementation Details.}
For all experiments on EPIC-KITCHENS-100 and SSv2-LT, we use Motionformer~\cite{Patrick2021}, 
a spatio-temporal transformer with attention guided by trajectories
which achieves strong results on EPIC-KITCHENS-100 and SSv2. We use the default configuration of 16 frame input and 224$\times$224 resolution with 16$\times$16 patches.
We train on 8$\times$V100 GPUs, with a distributed batch of 56 samples. To enable processing on multiple GPUs, we maintain a feature bank of previous iterations per GPU. 
Other details (architecture, optimisation \etc) follow the default code of Motionformer and are noted in Appendix \ref{app_motionformer}.
For all methods apart from CE and EQL, we follow the cRT disentanglement approach~\cite{Kang2020}. We first train end-to-end using instance-balanced sampling with a cross-entropy loss. We then reset the classifier and switch to class-balanced sampling for a full training run. 

For VideoLT-LT experiments, we use the codebase provided with the original dataset and accompanying method Framestack~\cite{Zhang2021} to be directly comparable to prior works. It uses pre-extracted ResNet-50 \cite{He2016} frame features with a non-linear classifier and score aggregation. 
We use the default batch size of 128 samples trained on 1$\times$P100 GPU.

For \method, the few-shot threshold is $\omega=20$. Decay and scaling parameters for the contribution function are $d=0.25$ and $l=0.6$ for SSv2-LT and VideoLT-LT, and $d=0.15$ and $l=1.0$ for EPIC-KITCHENS-100 as it has a smaller minimum class size.

\noindent\textbf{Results.}
Table \ref{tab:results_mf} shows the results for EPIC-KITCHENS-100, SSv2-LT and VideoLT-LT. \method\ performs best on all datasets for average class accuracy. 
Note that prior results were reported on datasets that did not contain any few shot classes (see Sec~\ref{sec:lt_image_datasets}).
By evaluating on EPIC-KITCHENS-100, and proposing benchmarks with few-shot classes, we can expose the limitations of these methods previously deemed competitive for long-tail video recognition.
\method\ also obtains the best results on few-shot classes (highlighted in green) on all datasets.
For tail classes, \method\ performs comparably or outperforms prior baselines.
For head classes, \method\ performs comparably to long-tail baselines on EPIC-KITCHENS-100 and SSv2-LT, but takes a bigger hit on VideoLT-LT.
We do not change any of the hyperparameters across datasets for fairer comparison, but consider results can be further improved if optimised per dataset.

Figure~\ref{fig:epic_class_improvements} shows class improvements of \method\ compared to CE on EPIC-KITCHENS-100. Significant improvements are seen on smaller classes (few-shot and end of tail). Some head classes drop in performance, particularly the largest. Similar trends were found on SSv2-LT and VideoLT-LT. 

\begin{figure}[t]
\centering
\includegraphics[width=1\columnwidth,trim={5 20 0 0}]{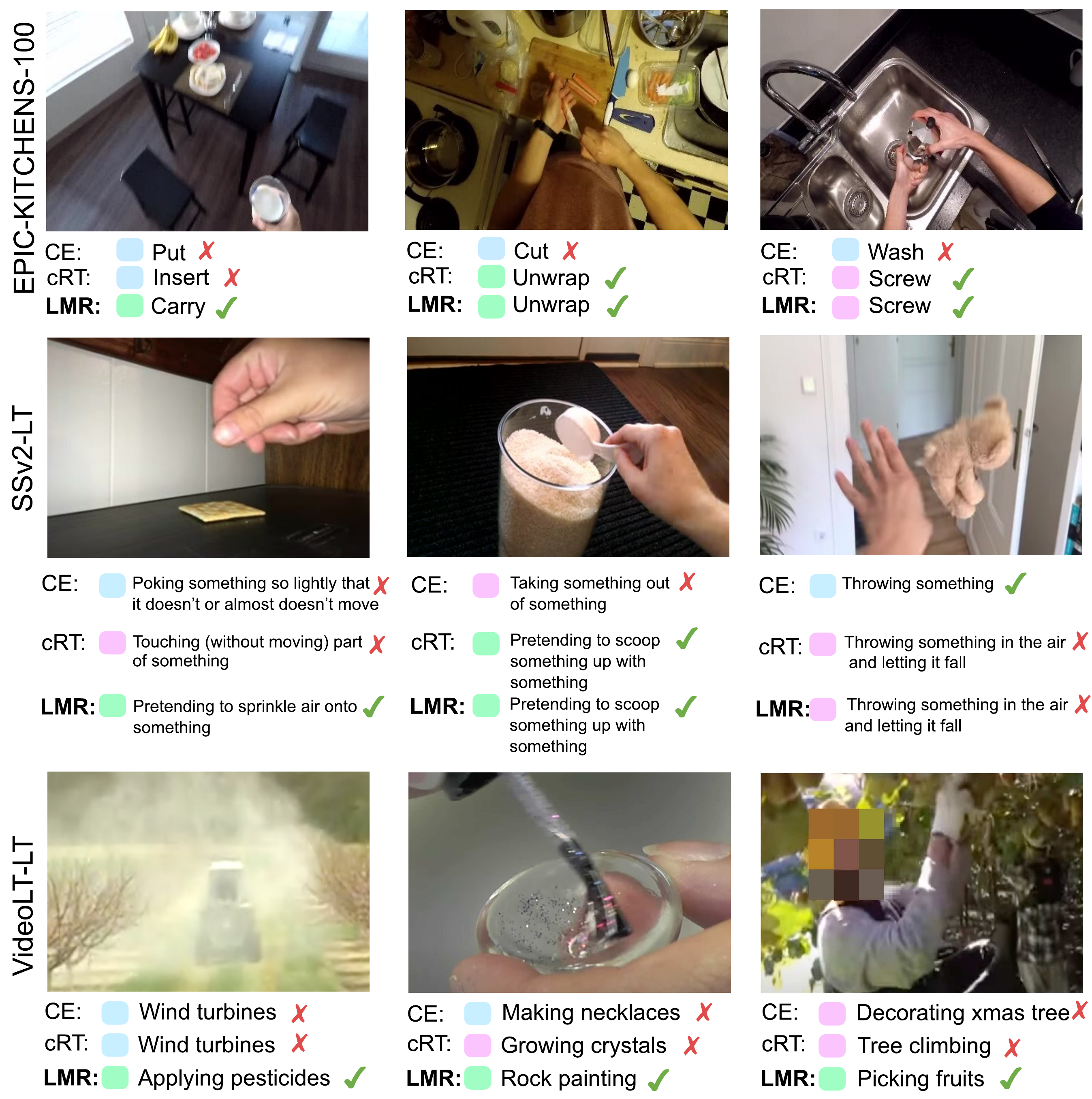}
\caption{Qualitative examples from all benchmarks comparing CE, cRT and the proposed \method. Blue, pink and green indicate whether the prediction is from a head, tail or few-shot class.
}
\label{fig:qual}
\vspace{-10pt}
\end{figure}

Figure~\ref{fig:qual} shows selected examples from all datasets. CE tends to predict few-shot classes as visually similar head classes. For example, on EPIC-KITCHENS-100, CE misclassifies the few-shot ``carry'' as the head class ``put'' due to visual similarity of holding the cup. Consistently, \method\ predicts the few-shot class correctly. A failure case is shown for SSv2-LT, where \method\ predicts the head class ``throwing something'' as the tail class ``throwing something in the air and letting it fall.''

\section{Ablations}\label{sec:ablation_method}

We perform all ablations on EPIC-KITCHENS-100 and SSv2-LT using the Motionformer backbone.

\noindent \textbf{\method\ Ablation. }
Table \ref{tab:method_ablation} ablates the design choices of \method\ against the full version (first row). First, class contributions are replaced by a constant (0.5 in A and 1 in B). When reconstructions are used solely, without the residual connection (B), performance decreases dramatically.
Using label mixing without reconstructions is shown in (C) as well as reconstructions without label mixing (D).
Interestingly, label mixing has a bigger impact on performance for SSv2-LT than EPIC-KITCHENS-100.

\begin{table*}[]
\centering
\resizebox{0.93\textwidth}{!}{
\footnotesize
\begin{tabular}{@{}lrrrcrrrc@{}}
\toprule
                                                                                                     & \multicolumn{4}{c}{\textbf{EPIC-KITCHENS-100}}                                                                                                                                                   & \multicolumn{4}{c}{\textbf{SSv2-LT}}                                                                                                                                                \\ 
Method Variant                                                                                       & \multicolumn{1}{l}{\cellcolor[HTML]{D7FFDA}Few} & \multicolumn{1}{l}{\cellcolor[HTML]{FCD7FF}Tail} & \multicolumn{1}{l}{\cellcolor[HTML]{D7EBFF}Head} & \multicolumn{1}{l}{Avg C/A} & \multicolumn{1}{l}{\cellcolor[HTML]{D7FFDA}Few} & \multicolumn{1}{l}{\cellcolor[HTML]{FCD7FF}Tail} & \multicolumn{1}{l}{\cellcolor[HTML]{D7EBFF}Head} & \multicolumn{1}{l}{Avg C/A} \\ \midrule
LMR                                                                                                  & \cellcolor[HTML]{D7FFDA}\textbf{35.7}           & \cellcolor[HTML]{FCD7FF}36.8                     & \cellcolor[HTML]{D7EBFF}51.1                     & \textbf{39.7}               & \cellcolor[HTML]{D7FFDA}17.9                    & \cellcolor[HTML]{FCD7FF}\textbf{46.5}            & \cellcolor[HTML]{D7EBFF}61.0                     & \textbf{38.3}               \\
(A) Constant contribution [replace Eq. \ref{eq:contribution} with $c(y) = 0.5$]                      & \cellcolor[HTML]{D7FFDA}34.1                    & \cellcolor[HTML]{FCD7FF}\textbf{37.1}            & \cellcolor[HTML]{D7EBFF}49.3                     & 39.3                        & \cellcolor[HTML]{D7FFDA}16.8                    & \cellcolor[HTML]{FCD7FF}44.9                     & \cellcolor[HTML]{D7EBFF}\textbf{61.9}            & 37.1                        \\
(B) No original representation in reconstruction [replace Eq. \ref{eq:contribution} with $c(y) = 1$] & \cellcolor[HTML]{D7FFDA}4.5                     & \cellcolor[HTML]{FCD7FF}2.5                      & \cellcolor[HTML]{D7EBFF}5.0                      & 3.4                         & \cellcolor[HTML]{D7FFDA}4.8                     & \cellcolor[HTML]{FCD7FF}7.4                      & \cellcolor[HTML]{D7EBFF}18.1                     & 6.0                         \\
(C) No reconstruction [replace Eq. \ref{eq:contribution_rep_rec} with $R = Z$]                       & \cellcolor[HTML]{D7FFDA}20.2                    & \cellcolor[HTML]{FCD7FF}36.7                     & \cellcolor[HTML]{D7EBFF}52.0                     & 38.1                        & \cellcolor[HTML]{D7FFDA}16.7                    & \cellcolor[HTML]{FCD7FF}46.4                     & \cellcolor[HTML]{D7EBFF}59.4                     & 37.6                        \\
(D) No pairwise label mixing [replace Eq. \ref{eq:r_final} with $\mathbf{mr}(Z,Y) = (R,Y)$]                       & \cellcolor[HTML]{D7FFDA}24.6                    & \cellcolor[HTML]{FCD7FF}33.9                     & \cellcolor[HTML]{D7EBFF}\textbf{53.2}            & 37.1                        & \cellcolor[HTML]{D7FFDA}\textbf{18.0}           & \cellcolor[HTML]{FCD7FF}45.9                     & \cellcolor[HTML]{D7EBFF}59.0                     & 37.9                        \\  \bottomrule
\end{tabular}
}
\vspace{-3pt}
\caption{Ablating \method\ on EPIC-KITCHENS-100 and SSv2-LT.\label{tab:method_ablation}}
\vspace{-10pt}
\end{table*}

\begin{table}
\vspace{-10pt}
\begin{minipage}{.225\textwidth}
\resizebox{\columnwidth}{!}{
\centering
\footnotesize
\begin{tabular}{@{}r
>{\columncolor[HTML]{D7FFDA}}r 
>{\columncolor[HTML]{FCD7FF}}r 
>{\columncolor[HTML]{D7EBFF}}r c@{}}
\toprule
\multicolumn{1}{l}{$l$} & \multicolumn{1}{l}{\cellcolor[HTML]{D7FFDA}Few} & \multicolumn{1}{l}{\cellcolor[HTML]{FCD7FF}Tail} & \multicolumn{1}{l}{\cellcolor[HTML]{D7EBFF}Head} & \multicolumn{1}{l}{Acc} \\ \midrule
0.0                       & 16.7                                            & 46.4                                             & 59.4                                             & 37.6                        \\
0.2                     & 16.9                                            & 46.3                                             & 59.0                                             & 37.7                        \\
0.4                     & 16.9                                            & 46.0                                             & 58.6                                             & 37.6                        \\
0.6                     & \textbf{17.9}                                   & 46.5                                             & 61.0                                             & \textbf{38.3}               \\
0.8                     & 17.6                                            & 46.5                                             & \textbf{61.9}                                    & 38.2                        \\
1.0                       & 16.2                                            & \textbf{46.7}                                    & 60.5                                             & 37.7                        \\ \bottomrule
\end{tabular}
}
\vspace{-4pt}
\caption{Effect of changing $l$, the contribution applied to the lowest class count on SSv2-LT. A higher $l$ means reconstructions contribute more to the representations. \label{tab:contribution_l}}
\end{minipage}%
\hfill
\begin{minipage}{0.23\textwidth}
\vspace{0pt}
\resizebox{\columnwidth}{!}{
\centering
\footnotesize
\begin{tabular}{@{}l
>{\columncolor[HTML]{D7FFDA}}r 
>{\columncolor[HTML]{FCD7FF}}r 
>{\columncolor[HTML]{D7EBFF}}r c@{}}
\toprule
\multicolumn{1}{l}{$d$} & \multicolumn{1}{l}{\cellcolor[HTML]{D7FFDA}Few} & \multicolumn{1}{l}{\cellcolor[HTML]{FCD7FF}Tail} & \multicolumn{1}{l}{\cellcolor[HTML]{D7EBFF}Head} & \multicolumn{1}{l}{Acc} \\ \midrule
0.11                    & 17.7                                            & 46.1                                             & 60.0                                             & 38.0                        \\
0.25                    & \textbf{17.9}                                   & 46.5                                             & \textbf{61.0}                                    & \textbf{38.3}               \\
0.5                     & 13.7                                            & 47.5                                             & 60.0                                             & 37.5                        \\
1.0                       & 12.4                                            & \textbf{48.0}                                    & 59.5                                             & 37.4                        \\ \bottomrule
\end{tabular}
}
\vspace{-4pt}
\caption{Effect of changing $d$, the decay of the class count contribution, on SSv2-LT. A lower $d$ means the contributions of reconstructions decay faster as class counts increase. \label{tab:contribution_d}}
\vspace{4.5pt}
\end{minipage}%
\vspace{-5pt}
\end{table}

\begin{table}
\begin{minipage}{0.23\textwidth}
\resizebox{\columnwidth}{!}{
\centering
\footnotesize
\begin{tabular}{@{}r 
>{\columncolor[HTML]{D7FFDA}}r 
>{\columncolor[HTML]{FCD7FF}}l 
>{\columncolor[HTML]{D7EBFF}}r c@{}} \toprule
\multicolumn{1}{l}{B} & \multicolumn{1}{l}{\cellcolor[HTML]{D7FFDA}Few} & Tail                                                      & \multicolumn{1}{l}{\cellcolor[HTML]{D7EBFF}Head} & \multicolumn{1}{l}{Acc} \\ \midrule
14                              & 17.1                                            & \multicolumn{1}{r}{\cellcolor[HTML]{FCD7FF}\textbf{46.7}} & 60.0                                             & 38.0                        \\
56                              & \textbf{17.9}                                   & \multicolumn{1}{r}{\cellcolor[HTML]{FCD7FF}46.5}          & \textbf{61.0}                                    & \textbf{38.3}               \\
224                             & 17.6                                            & 46.2                                                      & 60.0                                             & 38.0                        \\
896                             & 17.6                                            & 46.0                                                      & 59.0                                             & 37.9                       \\ \bottomrule
\end{tabular}
}
\vspace{-4pt}
\caption{Effect of varying the number of samples ($B$) used for reconstruction on SSv2-LT.  \label{tab:reconstruction_size}}
\vspace{11pt}
\end{minipage}%
\hfill
\begin{minipage}{0.23\textwidth}
\resizebox{\columnwidth}{!}{
\centering
\footnotesize
\begin{tabular}{@{}r
>{\columncolor[HTML]{D7FFDA}}r 
>{\columncolor[HTML]{FCD7FF}}r 
>{\columncolor[HTML]{D7EBFF}}r c@{}}
\toprule
\multicolumn{1}{l}{$\omega$} & \multicolumn{1}{l}{\cellcolor[HTML]{D7FFDA}Few} & \multicolumn{1}{l}{\cellcolor[HTML]{FCD7FF}Tail} & \multicolumn{1}{l}{\cellcolor[HTML]{D7EBFF}Head} & \multicolumn{1}{l}{Acc} \\ \midrule
0                            & \textbf{18.0}                                   & 45.9                                             & 59.5                                             & 37.9                        \\
20                           & 17.9                                            & \textbf{46.5}                                    & \textbf{61.0}                                    & \textbf{38.3}               \\
50                           & 17.6                                            & 46.2                                             & 59.5                                             & 37.9                        \\
500                          & 17.5                                            & 46.2                                             & 59.0                                             & 37.9                        \\ \bottomrule
\end{tabular}
}
\vspace{-4pt}
\caption{Effect of changing $\omega$ on SSv2-LT, which is the minimum class size threshold for the reconstruction mask. \label{tab:fewshot_threshold}}
\end{minipage}%
\vspace{-10pt}
\end{table}

\noindent \textbf{Contribution parameters. }
Reconstructions are combined with original representations according to the contribution function $\mathbf{c(Y)}$ in Eq.~\ref{eq:contribution_rep_rec}, which maps class count to a contribution between 0 and 1. It is parameterised by the decay $d$ and the contribution $l$ for the lowest class count. First, $d$ is fixed at $0.25$ and $l$ is varied between 0.0 and 1.0. Results are shown in \cref{tab:contribution_l}, where 0.6 performs best on the few-shot classes and overall. Next, $l$ is fixed at 0.6 and $d$ is varied, with results shown in \cref{tab:contribution_d}. In both cases, results have a region of stability, with the best combination being $l=0.6$ and $d=0.25$.

\noindent\textbf{Number of Samples Used for Reconstruction. }
We assess the impact of the number of samples in the batch used in the reconstruction process ($B$). 
Table \ref{tab:reconstruction_size} shows how varying the number of samples affects overall performance on SSv2-LT. 
Best performance is reported at our default of 56 samples.

\noindent \textbf{Threshold for Masked Classes in Reconstruction. }
The threshold $\omega$, used for masking in Eq.~\ref{eq:mask}, is by default set to~20, which is the threshold for few-shot classes. The masking is used to prevent few-shot samples contributing to the reconstruction of other samples. Table \ref{tab:fewshot_threshold} shows the effect of varying $\omega$. 
Best performance is obtained at $\omega = 20$.

\begin{figure}
\centering
\subfloat{\includegraphics[width=0.6\columnwidth,trim={0 0 0 0 }]{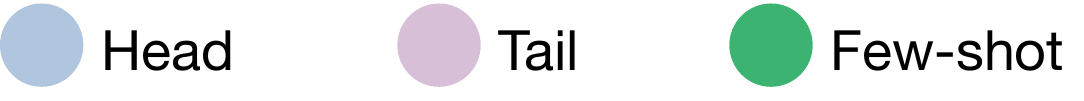}} \\
\addtocounter{subfigure}{-1}
\subfloat[EPIC-KITCHENS-100 cRT. \label{fig:proj_epic_no_reconstruction}]{\includegraphics[width=0.5\columnwidth,trim={190 190 170 170 },clip]{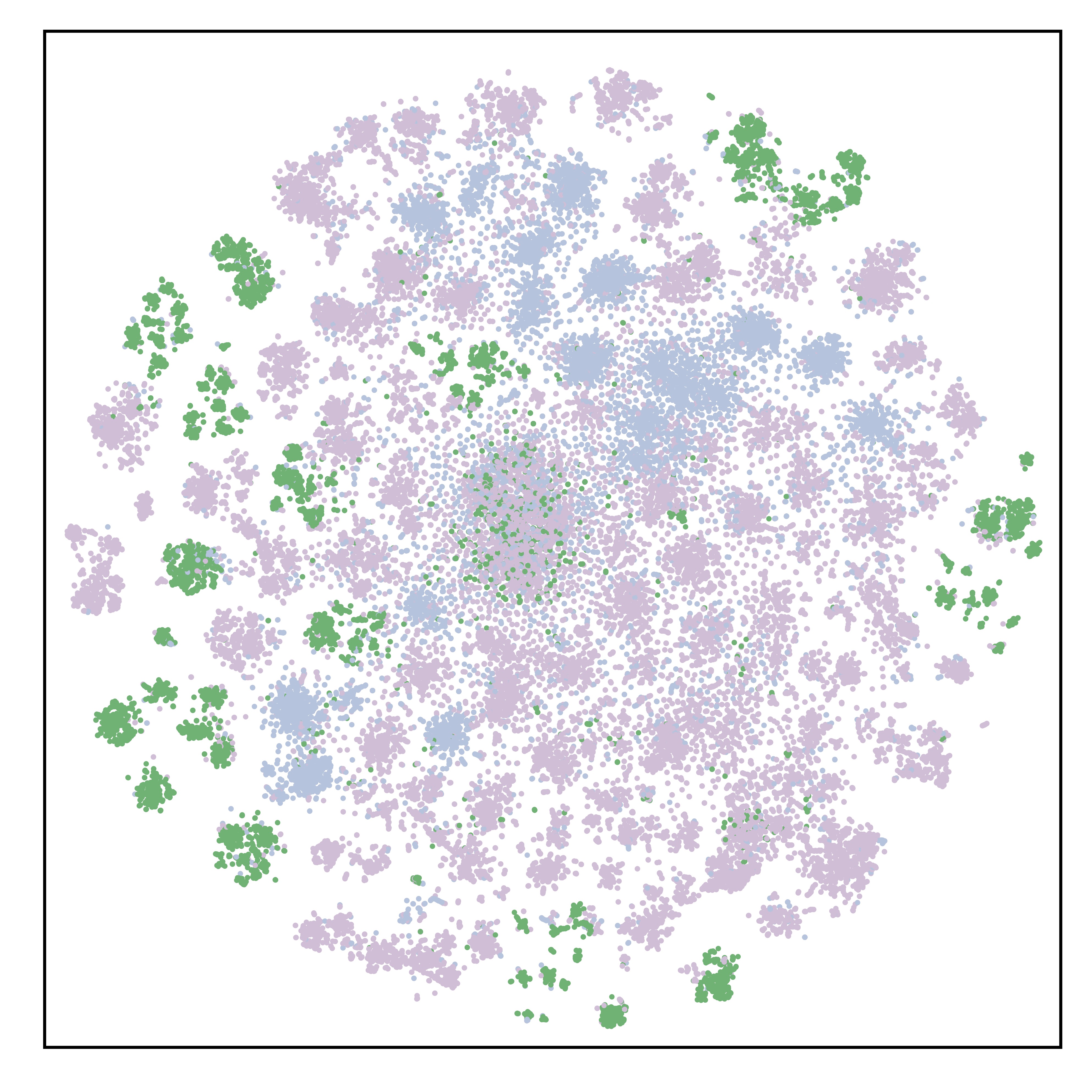}} 
\subfloat[EPIC-KITCHENS-100 \method. \label{fig:proj_epic_reconstruction}]{\includegraphics[width=0.5\columnwidth,trim={190 190 170 170 },clip]{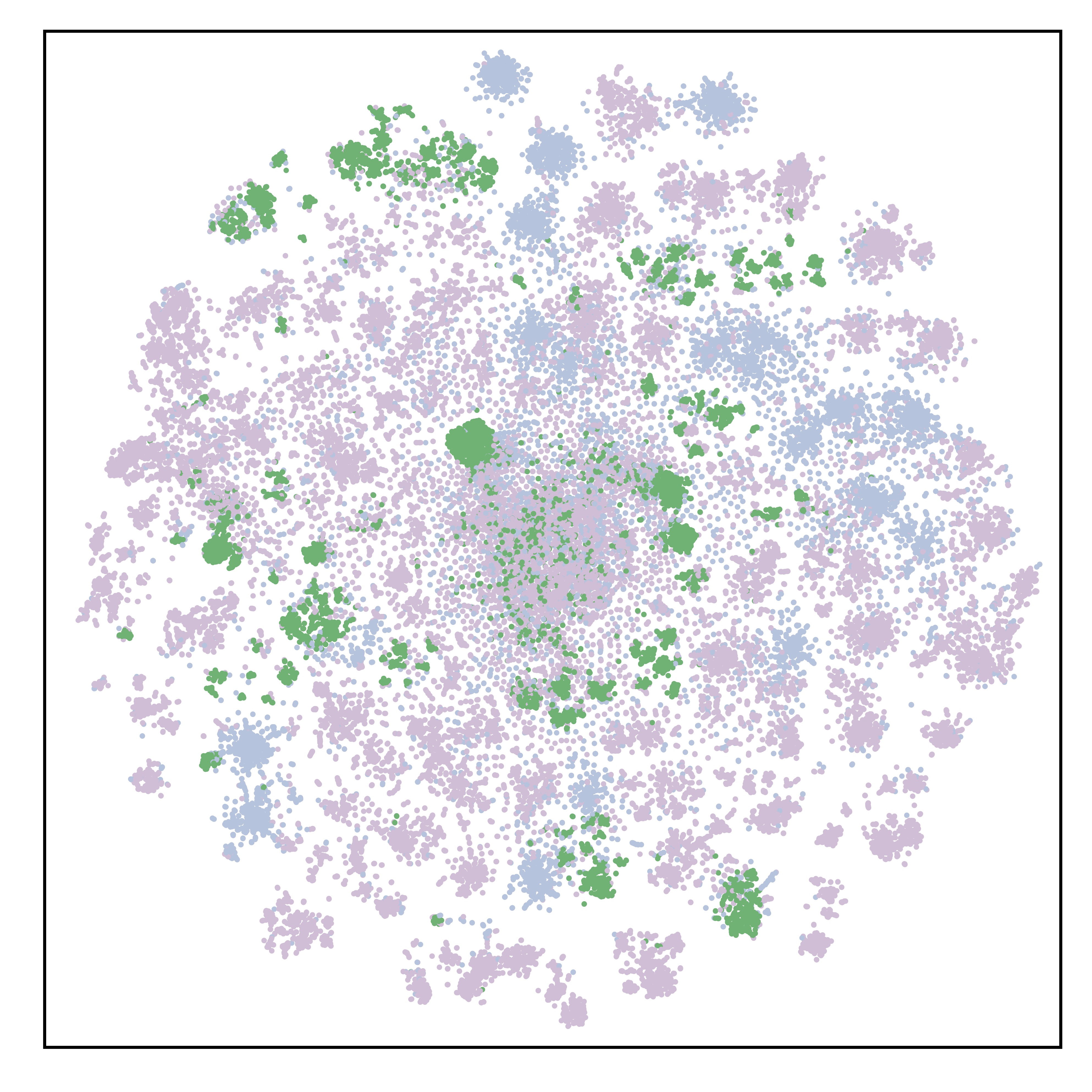}} \\
\subfloat[SSv2-LT cRT. \label{fig:proj_ssv2lt_no_reconstruction}]{\includegraphics[width=0.5\columnwidth,trim={190 190 170 170 },clip]{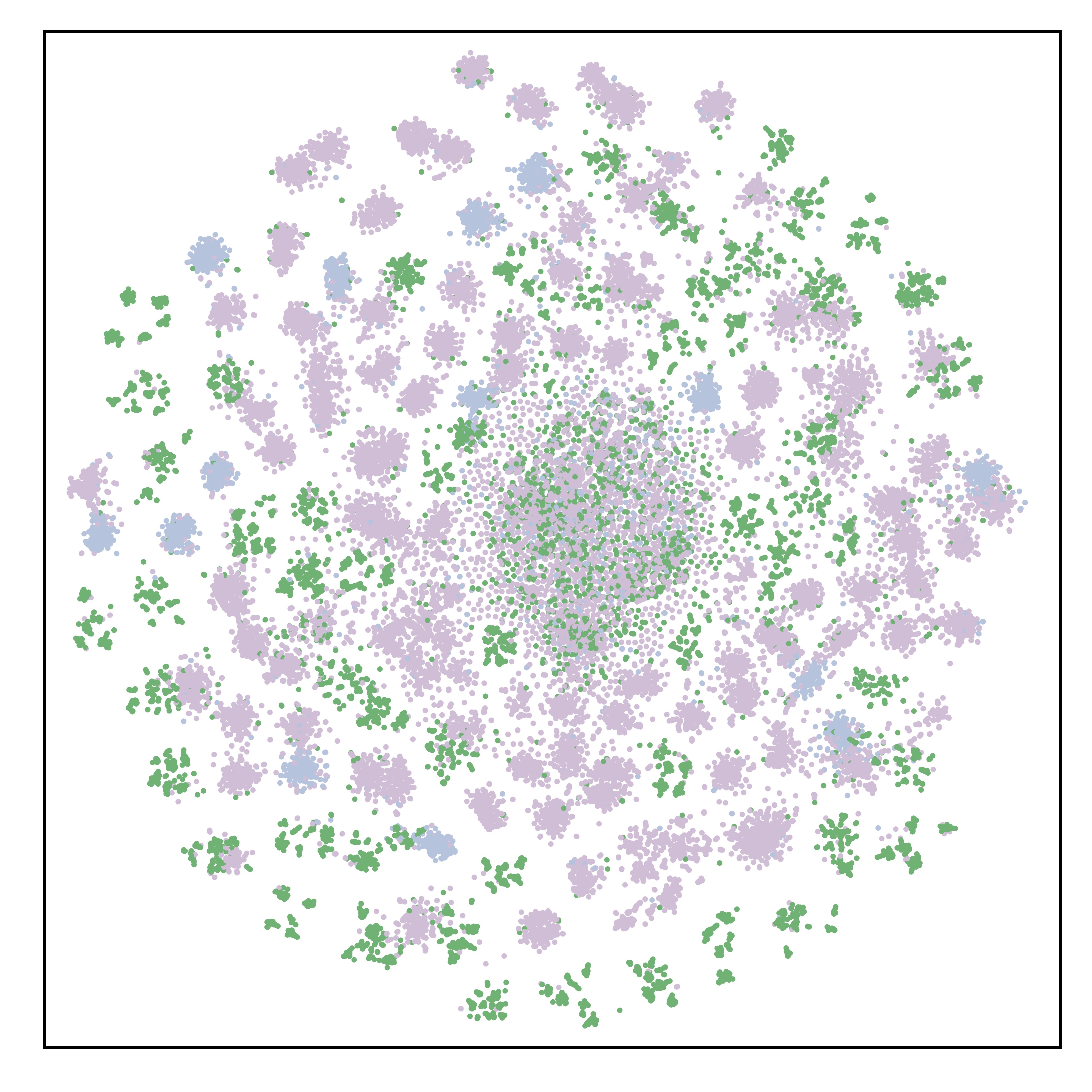}} 
\subfloat[SSv2-LT \method. \label{fig:proj_ssv2lt_reconstruction}]{\includegraphics[width=0.5\columnwidth,trim={190 190 170 170 },clip]{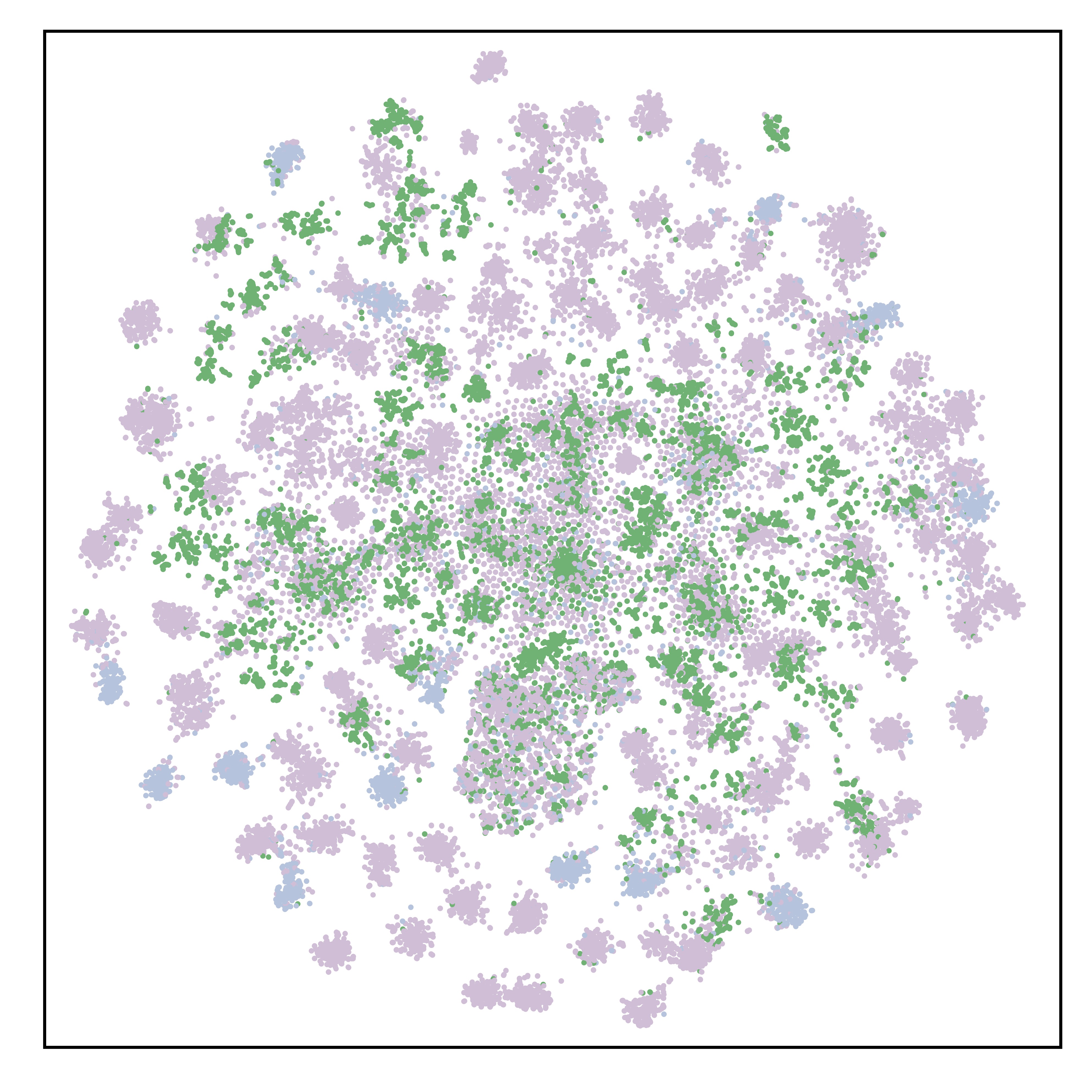}}
\vspace{-3pt}
\caption{Effect of \method\ on EPIC-KITCHENS-100 (top) and SSv2-LT (bottom) t-SNE projections. 
Without reconstruction (left), samples from few-shot classes (green) are pushed to the edge, and tightly clustered.
With \method\ (right), the few-shot clusters are larger and closer to the centre, i.e. in closer proximity to head (blue) and tail (pink) classes. This gives more robust boundaries as they are bordering more classes.
}
\vspace{-9pt}
\label{fig:projection}
\end{figure}

\noindent \textbf{Visualising \method\ . }
\cref{fig:projection} shows t-SNE \cite{van2008visualizing} projections of representations without \method\ (\ie cRT) and with.
cRT pushes the few shot classes (green) to the periphery.
\method\ results in larger, \ie more diverse, few-shot clusters towards the centre of the projection. This indicates a higher proximity to head and tail classes
 which creates robust class boundaries and better generality to unseen test samples.

\section{Conclusion}

In this paper, we defined a set of properties, enabling  quantitative comparison of long-tail distributions. We showcased that curated long-tail image datasets are comparable to naturally-collected ones, while previously proposed video datasets fall short. 
Based on these findings, we proposed new benchmarks, SSv2-LT and VideoLT-LT, and suggested their use, alongside EPIC-KITCHENS-100, for evaluating long-tail video recognition. 

We proposed 
LMR, a method for long-tail video recognition, which reconstructs few-shot samples as weighted combinations of  other samples in the batch.
A residual connection, weighted by the class size, combines instances with their reconstructions, followed by pairwise label mixing.
\method\ reduces overfitting to instances from few-shot classes, and outperforms prior methods on the three benchmarks. 

We hope our proposed benchmarks and method will provide a foundation for long-tail video recognition, and encourage further contributions applicable to naturally-collected data.

\noindent \textbf{Acknowledgments.} 
We use publicly available datasets and publish our proposed benchmarks.
Research is funded by EPSRC UMPIRE (EP/T004991/1), EPSRC SPHERE Next Steps (EP/R005273/1), EPSRC DTP and EPSRC PG Visual AI (EP/T028572/1). We acknowledge the use of HPC Tier~2 Facility Jade 2 and Bristol's Blue Crystal 4 facility.

{\small
\bibliographystyle{ieee_fullname}
\bibliography{egbib} 

\begin{thebibliography}{10}\itemsep=-1pt

\bibitem{Abu-El-Haija2016}
Sami Abu-El-Haija, Nisarg Kothari, Joonseok Lee, Paul Natsev, George Toderici,
  Balakrishnan Varadarajan, and Sudheendra Vijayanarasimhan.
\newblock {YouTube-8M: A Large-Scale Video Classification Benchmark}.
\newblock In {\em arXiv}, 2016.

\bibitem{Alayrac2022}
Jean-Baptiste Alayrac, Jeff Donahue, Pauline Luc, Antoine Miech, Iain Barr,
  Yana Hasson, Karel Lenc, Arthur Mensch, Katie Millican, Malcolm Reynolds,
  Roman Ring, Eliza Rutherford, Serkan Cabi, Tengda Han, Zhitao Gong, Sina
  Samangooei, Marianne Monteiro, Jacob Menick, Sebastian Borgeaud, Andrew
  Brock, Aida Nematzadeh, Sahand Sharifzadeh, Mikolaj Binkowski, Ricardo
  Barreira, Oriol Vinyals, Andrew Zisserman, and Karen Simonyan.
\newblock {Flamingo: a Visual Language Model for Few-Shot Learning}.
\newblock In {\em Advances in Nueral Information Processing Systems}, 2022.

\bibitem{weight_balancing}
Shaden Alshammari, Yu-Xiong Wang, Deva Ramanan, and Shu Kong.
\newblock Long-tailed recognition via weight balancing.
\newblock In {\em CVPR}, 2022.

\bibitem{Anderson}
Chris Anderson.
\newblock {\em {The Long Tail: Why the Future of Business Is Selling Less of
  More}}.
\newblock Hachette Books, 2006.

\bibitem{Bishay2019}
Mina Bishay, Georgios Zoumpourlis, and Ioannis Patras.
\newblock {TARN: Temporal Attentive Relation Network for Few-Shot and Zero-Shot
  Action Recognition}.
\newblock In {\em BMVC}, 2019.

\bibitem{Buda2018}
Mateusz Buda, Atsuto Maki, and Maciej~A Mazurowski.
\newblock {A systematic study of the class imbalance problem in convolutional
  neural networks}.
\newblock {\em Neural Networks}, (106):249--259, 2018.

\bibitem{Cai2021}
Jiarui Cai, Yizhou Wang, and Jenq-Neng Hwang.
\newblock {ACE: Ally Complementary Experts for Solving Long-Tailed Recognition
  in One-Shot}.
\newblock In {\em ICCV}, 2021.

\bibitem{Cao2020}
Kaidi Cao, Jingwei Ji, Zhangjie Cao, Chien-Yi Chang, and Juan~Carlos Niebles.
\newblock {Few-Shot Video Classification via Temporal Alignment}.
\newblock In {\em CVPR}, 2020.

\bibitem{Cao2019a}
Kaidi Cao, Colin Wei, Adrien Gaidon, Nikos Arechiga, and Tengyu Ma.
\newblock {Learning Imbalanced Datasets with Label-Distribution-Aware Margin
  Loss}.
\newblock In {\em NeurIPS}, 2019.

\bibitem{Carreira}
Joao Carreira and Andrew Zisserman.
\newblock {Quo Vadis, Action Recognition? A New Model and the Kinetics
  Dataset}.
\newblock {\em CVPR}, 2017.

\bibitem{imagine_by_reasoning}
Xiaohua Chen, Yucan Zhou, Dayan Wu, Wanqian Zhang, Yu Zhou, Bo Li, and Weiping
  Wang.
\newblock Imagine by reasoning: A reasoning-based implicit semantic data
  augmentation for long-tailed classification.
\newblock In {\em AAAI}, 2022.

\bibitem{Chu}
Peng Chu, Xiao Bian, Shaopeng Liu, and Haibin Ling.
\newblock {Feature Space Augmentation for Long-Tailed Data}.
\newblock In {\em ECCV}, 2020.

\bibitem{Cui2021}
Jiequan Cui, Shu Liu, Zhuotao Tian, Zhisheng Zhong, and Jiaya Jia.
\newblock Reslt: Residual learning for long-tailed recognition.
\newblock {\em IEEE TPAMI}, 2022.

\bibitem{Cuia}
Jiequan Cui, Zhisheng Zhong, Shu Liu, Yu Bei, and Jia Jiaya.
\newblock {Parametric Contrastive Learning}.
\newblock In {\em ICCV}, 2021.

\bibitem{Cui2019}
Yin Cui, Menglin Jia, Tsung~Yi Lin, Yang Song, and Serge Belongie.
\newblock {Class-Balanced Loss Based on Effective Number of Samples}.
\newblock {\em CVPR}, 2019.

\bibitem{Damen2021}
Dima Damen, Hazel Doughty, Giovanni~Maria Farinella, Antonino Furnari,
  Evangelos Kazakos, Jian Ma, Davide Moltisanti, Jonathan Munro, Toby Perrett,
  Will Price, and Michael Wray.
\newblock {Rescaling Egocentric Vision}.
\newblock {\em IJCV}, 2021.

\bibitem{Deng2009}
Jia Deng, Wei Dong, Richard Socher, Li-Jia Li, Kai Li, and Li Fei-Fei.
\newblock {ImageNet: A Large-Scale Hierarchical Image Database}.
\newblock In {\em CVPR}, 2009.

\bibitem{Wangd}
Zongyong Deng, Hao Liu, Yaoxing Wang, Chenyang Wang, Zekuan Yu, and Xuehong
  Sun.
\newblock {PML : Progressive Margin Loss for Long-tailed Age Classification}.
\newblock In {\em CVPR}, 2021.

\bibitem{Doersch2020}
Carl Doersch, Ankush Gupta, and Andrew Zisserman.
\newblock {CrossTransformers: Spatially-Aware Few-Shot Transfer}.
\newblock In {\em NeurIPS}, 2020.

\bibitem{Dwivedi2019}
Sai~Kumar Dwivedi, Vikram Gupta, Rahul Mitra, Shuaib Ahmed, and Arjun Jain.
\newblock {ProtoGAN: Towards Few Shot Learning for Action Recognition}.
\newblock In {\em CVPR}, 2019.

\bibitem{Goyal2017}
Raghav Goyal, Vincent Michalski, Joanna Materzy, Susanne Westphal, Heuna Kim,
  Valentin Haenel, Peter Yianilos, Moritz Mueller-freitag, Florian Hoppe,
  Christian Thurau, Ingo Bax, and Roland Memisevic.
\newblock {The “Something Something” Video Database for Learning and
  Evaluating Visual Common Sense}.
\newblock In {\em ICCV}, 2017.

\bibitem{LVIS}
Agrim Gupta, Piotr Dollár, and Ross Girshick.
\newblock Lvis: A dataset for large vocabulary instance segmentation.
\newblock In {\em CVPR}, 2019.

\bibitem{He2016}
Kaiming He, Xiangyu Zhang, Shaoqing Ren, and Jian Sun.
\newblock {Deep Residual Learning for Image Recognition}.
\newblock In {\em CVPR}, 2016.

\bibitem{Horn2018}
Grant~Van Horn, Oisin~Mac Aodha, Yang Song, Yin Cui, Chen Sun, Alex Shepard,
  Hartwig Adam, Pietro Perona, and Serge Belongie.
\newblock {The iNaturalist Species Classification and Detection Dataset}.
\newblock In {\em CVPR}, 2018.

\bibitem{jamal_2020}
Muhammad~Abdullah Jamal, Matthew Brown, Ming-Hsuan Yang, Liqiang Wang, and
  Boqing Gong.
\newblock Rethinking class-balanced methods for long-tailed visual recognition
  from a domain adaptation perspective.
\newblock In {\em CVPR}, 2020.

\bibitem{Kang2020}
Bingyi Kang, Saining Xie, Marcus Rohrbach, Zhicheng Yan, Albert Gordo, Jiashi
  Feng, and Yannis Kalantidis.
\newblock {Decoupling Representation and Classifier for Long-Tailed
  Recognition}.
\newblock In {\em ICLR}, 2020.

\bibitem{Kasarla2022}
Tejaswi Kasarla, Gertjan~J. Burghouts, Max van Spengler, Elise van~der Pol,
  Rita Cucchiara, and Pascal Mettes.
\newblock {Maximum Class Separation as Inductive Bias in One Matrix}.
\newblock In {\em NeurIPS}, 2022.

\bibitem{Kim}
Jaehyung Kim, Jongheon Jeong, and Jinwoo Shin.
\newblock {M2m: Imbalanced Classification via Major-to-minor Translation}.
\newblock In {\em CVPR}, 2020.

\bibitem{Krizhevsky2009}
Alex Krizhevsky.
\newblock {Learning Multiple Layers of Features from Tiny Images}.
\newblock 2009.

\bibitem{trustworthy_classification}
Bolian Li, Zongbo Han, Haining Li, Huazhu Fu, and Changqing Zhang.
\newblock Trustworthy long-tailed classification.
\newblock In {\em CVPR}, 2022.

\bibitem{equalized_focal}
Bo Li, Yongqiang Yao, Jingru Tan, Gang Zhang, Fengwei Yu, Jianwei Lu, and Ye
  Luo.
\newblock Equalized focal loss for dense long-tailed object detection.
\newblock In {\em CVPR}, 2022.

\bibitem{gaussian_la}
Mengke Li, Yiu-ming Cheung, and Yang Lu.
\newblock Long-tailed visual recognition via gaussian clouded logit adjustment.
\newblock In {\em CVPR}, 2022.

\bibitem{Lia}
Shuang Li, Kaixiong Gong, Chi Harold, Liu Yulin, Wang Feng, and Qiao Xinjing.
\newblock {MetaSAug : Meta Semantic Augmentation for Long-Tailed Visual
  Recognition}.
\newblock In {\em CVPR}, 2021.

\bibitem{TSC}
Tianhong Li, Peng Cao, Yuan Yuan, Lijie Fan, Yuzhe Yang, Rogerio~S. Feris,
  Piotr Indyk, and Dina Katabi.
\newblock Targeted supervised contrastive learning for long-tailed recognition.
\newblock In {\em CVPR}, 2022.

\bibitem{Lin}
Tsung-yi Lin, Priya Goyal, Ross Girshick, Kaiming He, and Piotr Dollar.
\newblock {Focal Loss for Dense Object Detection}.
\newblock In {\em ICCV}, 2017.

\bibitem{Liu2020a}
Jialun Liu, Yifan Sun, Chuchu Han, Zhaopeng Dou, and Wenhui Li.
\newblock {Deep representation learning on long-tailed data: A learnable
  embedding augmentation perspective}.
\newblock In {\em CVPR}, 2020.

\bibitem{Liu2019}
Ziwei Liu, Zhongqi Miao, Xiaohang Zhan, Jiayun Wang, Boqing Gong, and Stella~X.
  Yu.
\newblock Large-scale long-tailed recognition in an open world.
\newblock In {\em CVPR}, 2019.

\bibitem{OLTR_TPAMI}
Ziwei Liu, Zhongqi Miao, Xiaohang Zhan, Jiayun Wang, Boqing Gong, and Stella~X.
  Yu.
\newblock Open long-tailed recognition in a dynamic world.
\newblock {\em IEEE TPAMI}, 2022.

\bibitem{Patrick2021}
Patrick Mandela, Dylan Campbell, Yuki Asano, Ishan Misra, Florian Metze,
  Christoph Feichtenhofer, Andrea Vedaldi, and Jo{\~{a}}o~F. Henriques.
\newblock {Keeping Your Eye on the Ball : Trajectory Attention in Video
  Transformers}.
\newblock In {\em NeurIPS}, 2021.

\bibitem{Menon2021}
Aditya~Krishna Menon, Sadeep Jayanumana, Ankit~Singh Rawat, Himanshu Jain,
  Andreas Veit, and Sanjiv Kuma.
\newblock {Long-Tail Learning via Logit Adjustment}.
\newblock In {\em ICLR}, 2021.

\bibitem{Mishra}
Ashish Mishra, Vinay~Kumar Verma, M~Shiva~Krishna Reddy, S Arulkumar, Piyush
  Rai, and Anurag Mittal.
\newblock {A Generative Approach to Zero-Shot and Few-Shot Action Recognition}.
\newblock In {\em WACV}, 2018.

\bibitem{CMO}
S. Park, Y. Hong, B. Heo, S. Yun, and J. Choi.
\newblock The majority can help the minority: Context-rich minority
  oversampling for long-tailed classification.
\newblock In {\em CVPR}, 2022.

\bibitem{Park}
Seulki Park, Jongin Lim, Younghan Jeon, and Jin~Young Choi.
\newblock {Influence-Balanced Loss for Imbalanced Visual Classification}.
\newblock In {\em ICCV}, 2021.

\bibitem{Patrick2021a}
Mandela Patrick, Po-Yao Huang, Yuki Asano, Florian Metze, Alexander Hauptmann,
  Jo{\~{a}}o Henriques, and Andrea Vedaldi.
\newblock {Support-set bottlenecks for video-text representation learning}.
\newblock In {\em ICLR}, 2021.

\bibitem{Perrett2021}
Toby Perrett, Alessandro Masullo, Tilo Burghardt, Majid Mirmehdi, and Dima
  Damen.
\newblock {Temporal-Relational CrossTransformers for Few-Shot Action
  Recognition}.
\newblock In {\em CVPR}, 2021.

\bibitem{Ravi2017}
Sachin Ravi and Hugo Larochelle.
\newblock {Optimizaion as a Model for Few-Shot Learning}.
\newblock In {\em ICLR}, 2017.

\bibitem{Ren2020}
Jiawei Ren, Cunjun Yu, Shunan Sheng, Xiao Ma, Haiyu Zhao, Shuai Yi, and
  Hongsheng Li.
\newblock {Balanced Meta-Softmax for Long-Tailed Visual Recognition}.
\newblock In {\em NeurIPS}, 2020.

\bibitem{Upervised2018}
Mengye Ren, Eleni Triantafillou, Sachin Ravi, Jake Snell, Kevin Swersky,
  Joshua~B. Tenenbaum, Hugo Larochelle, and Richard~S. Zemel.
\newblock {Meta-Learning for Semi-Supervised Few-Shot Classification}.
\newblock In {\em ICLR}, 2018.

\bibitem{Samuel2021}
Dvir Samuel and Gal Chechik.
\newblock {Distributional Robustness Loss for Long-tail Learning}.
\newblock In {\em ICCV}, 2021.

\bibitem{meta-weight-net}
Jun Shu, Qi Xie, Lixuan Yi, Qian Zhao, Sanping Zhou, Zongben Xu, and Deyu Meng.
\newblock Meta-weight-net: Learning an explicit mapping for sample weighting.
\newblock In {\em NeurIPS}, 2019.

\bibitem{sinha_ijcv}
Saptarshi Sinha, Hiroki Ohashi, and Katsuyuki Nakamura.
\newblock {Class-Difficulty Based Methods for Long-Tailed Visual Recognition}.
\newblock {\em IJCV}, 2022.

\bibitem{Starr2008}
Susan Starr and Jeff Williams.
\newblock {The Long Tail: a Usage Analysis of Pre-1993 Print Biomedical Journal
  Literature}.
\newblock {\em Journal of the Medical Library Association}, 96(1), 2008.

\bibitem{EQL_v2}
Jingru Tan, Xin Lu, Gang Zhang, Changqing Yin, and Quanquan Li.
\newblock Equalization loss v2: A new gradient balance approach for long-tailed
  object detection.
\newblock In {\em CVPR}, 2021.

\bibitem{Tan2020}
Jingru Tan, Changbao Wang, Buyu Li, Quanquan Li, Wanli Ouyang, Changqing Yin,
  and Junjie Yan.
\newblock {Equalization loss for long-tailed object recognition}.
\newblock In {\em CVPR}, 2020.

\bibitem{invariant_feature_learning}
Kaihua Tang, Mingyuan Tao, Jiaxin Qi, Zhenguang Liu, and Hanwang Zhang.
\newblock Invariant feature learning for generalized long-tailed
  classification.
\newblock In {\em ECCV}, 2022.

\bibitem{Thatipelli2022}
Anirudh Thatipelli, Sanath Narayan, Salman Khan, Rao~Muhammad Anwer,
  Fahad~Shahbaz Khan, and Bernard Ghanem.
\newblock {Spatio-Temporal Relation Modeling for Few-shot Action Recognition}.
\newblock In {\em CVPR}, 2022.

\bibitem{Tian2020}
Junjiao Tian, Yen-Cheng Liu, Nathaniel Glaser, Yen-Chang Hsy, and Zsolt Kira.
\newblock {Posterior Re-calibration for Imbalanced Datasets}.
\newblock In {\em NeurIPS}, 2020.

\bibitem{Tokmakov2019}
Pavel Tokmakov, Yu-Xiong Wang, and Martial Hebert.
\newblock {Learning Compositional Representations for Few-Shot Recognition}.
\newblock In {\em ICCV}, 2019.

\bibitem{triantafillou2019metadataset}
Eleni Triantafillou, Tyler Zhu, Vincent Dumoulin, Pascal Lamblin, Utku Evci,
  Kelvin Xu, Ross Goroshin, Carles Gelada, Kevin Swersky, Pierre-Antoine
  Manzagol, and Hugo Larochelle.
\newblock {Meta-Dataset: A Dataset of Datasets for Learning to Learn from Few
  Examples}.
\newblock In {\em International Conference on Learning Representations}, 2020.

\bibitem{van2008visualizing}
Laurens Van~der Maaten and Geoffrey Hinton.
\newblock Visualizing data using t-sne.
\newblock {\em Journal of machine learning research}, 9(11), 2008.

\bibitem{Vinyals2016}
Oriol Vinyals, Charles Blundell, Timothy Lillicrap, Koray Kavukcuoglu, and Daan
  Wierstra.
\newblock {Matching Networks for One Shot Learning}.
\newblock In {\em NeurIPS}, 2016.

\bibitem{ride}
Xudong Wang, Long Lian, Zhongqi Miao, Ziwei Liu, and Stella Yu.
\newblock Long-tailed recognition by routing diverse distribution-aware
  experts.
\newblock In {\em ICLR}, 2021.

\bibitem{Wang2022}
Xiang Wang, Shiwei Zhang, Zhiwu Qing, Mingqian Tang, Zhengrong Zuo, Changxin
  Gao, Rong Jin, and Nong Sang.
\newblock Hybrid relation guided set matching for few-shot action recognition.
\newblock In {\em CVPR}, 2022.

\bibitem{Wang2017a}
Yu~Xiong Wang, Deva Ramanan, and Martial Hebert.
\newblock {Learning to model the tail}.
\newblock In {\em NeurIPS}, 2017.

\bibitem{Wu2022}
Chao-Yuan Wu, Yanghao Li, Karttikeya Mangalam, Haoqi Fan, Bo Xiong, Jitendra
  Malik, and Christoph Feichtenhofer.
\newblock {MeMViT: Memory-Augmented Multiscale Vision Transformer for Efficient
  Long-Term Video Recognition}.
\newblock In {\em CVPR}, 2022.

\bibitem{adversarial_lt}
Tong Wu, Ziwei Liu, Qingqiu Huang, Yu Wang, and Dahua Lin.
\newblock Adversarial robustness under long-tailed distribution.
\newblock In {\em CVPR}, 2021.

\bibitem{CVPR_rebalancing}
Sihao Yu, Jiafeng Guo, Ruqing Zhang, Yixing Fan, Zizhen Wang, and Xueqi Cheng.
\newblock A re-balancing strategy for class-imbalanced classification based on
  instance difficulty.
\newblock In {\em CVPR}, 2022.

\bibitem{Zhang2018a}
Hongyi Zhang, Moustapha Cisse, Yann~N. Dauphin, and David Lopez-Paz.
\newblock {mixup : Beyond Empirical Risk Minimization}.
\newblock In {\em ICLR}, 2018.

\bibitem{Zhang2020}
Hongguang Zhang, Li Zhang, Xiaojuan Qi, Hongdong Li, Philip H~S Torr, and Piotr
  Koniusz.
\newblock {Few-shot Action Recognition with Permutation-invariant Attention}.
\newblock In {\em ECCV}, 2020.

\bibitem{DisAlign}
Songyang Zhang, Zeming Li, Shipeng Yan, Xuming He, and Jian Sun.
\newblock Distribution alignment: A unified framework for long-tail visual
  recognition.
\newblock In {\em CVPR}, 2021.

\bibitem{Zhang2021}
Xing Zhang, Zuxuan Wu, Zejia Weng, Huazhu Fu, Jingjing Chen, Yu-Gang Jiang, and
  Larry Davis.
\newblock {VideoLT: Large-scale Long-tailed Video Recognition}.
\newblock In {\em ICCV}, 2021.

\bibitem{Zhangb}
Yifan Zhang, Bryan Hooi, Lanqing Hong, and Jiashi Feng.
\newblock {Test-Agnostic Long-Tailed Recognition by Test-Time Aggregating
  Diverse Experts with Self-Supervision}.
\newblock In {\em Advances in Nerual Information Processing Systems Workshop},
  2021.

\bibitem{MiSLAS}
Zhisheng Zhong, Jiequan Cui, Shu Liu, and Jiaya Jia.
\newblock Improving calibration for long-tailed recognition.
\newblock In {\em CVPR}, 2021.

\bibitem{balanced_contrastive}
Jianggang Zhu, Zheng Wang, Jingjing Chen, Yi-Ping~Phoebe Chen, and Yu-Gang
  Jiang.
\newblock Balanced contrastive learning for long-tailed visual recognition.
\newblock In {\em CVPR}, 2022.

\bibitem{Zhu2018}
Linchao Zhu and Yi Yang.
\newblock {Compound Memory Networks for Few-Shot Video Classification}.
\newblock In {\em ECCV}, 2018.

\bibitem{Zhu2020a}
Linchao Zhu and Yi Yang.
\newblock {Inflated episodic memory with region self-attention for long-tailed
  visual recognition}.
\newblock In {\em CVPR}, 2020.

\bibitem{Zhu2020}
Linchao Zhu and Yi Yang.
\newblock {Label Independent Memory for Semi-Supervised Few-shot Video
  Classification}.
\newblock {\em IEEE TPAMI}, 14(8), 2020.

\bibitem{Zhu2021}
Xiatian Zhu, Antoine Toisoul, Juan-Manuel P{\'{e}}rez-R{\'{u}}a, Li Zhang,
  Brais Martinez, and Tao Xiang.
\newblock {Few-shot Action Recognition with Prototype-centered Attentive
  Learning}.
\newblock In {\em BMVC}, 2021.

\end{thebibliography}
}

\appendix

\section{Video-LT and SSv2-LT Datasets}\label{app_datasets}

In Section \ref{sec:proposed_datasets}, we curated long-tail versions of SSv2 \cite{Goyal2017} and VideoLT \cite{Zhang2021}. More details are provided here.

\noindent \textbf{SSv2-LT:} We follow the recipe used for ImageNet-LT and Places-LT from~\cite{Liu2019} and use the Pareto distribution 
with $\alpha=6$ and a minimum class count of 5. We rank classes by their original size in the training set (\ie the largest class in SSv2 is still the largest class in SSv2-LT and so on). We take a maximum class size of 2500, which is as large as it can be given the original and curated dataset sizes. 
Balanced training and validation sets are taken from the original training split, and the test set is taken from the original validation split (labels are not available for the test split from \cite{Goyal2017}). 

\noindent \textbf{VideoLT-LT:} We use the same recipe as above, setting the maximum class size of 550, and keep the minimum as 5 and $\alpha$ as 6. 
We sample the proposed long-tail train split from the original VideoLT train split, and sample balanced val and test sets from the original unbalanced val and test test splits respectively.
We do not include test videos with multiple labels (around 10\%), and we do not include classes with fewer than 10 test samples. This maintains 772 classes, and ensures our smallest classes are evaluated robustly.

\begin{figure}
\centering
\subfloat[Something-Something V2.\label{fig:ssv2construction}]{\includegraphics[width=0.5\linewidth,trim={0 1mm 0 0}]{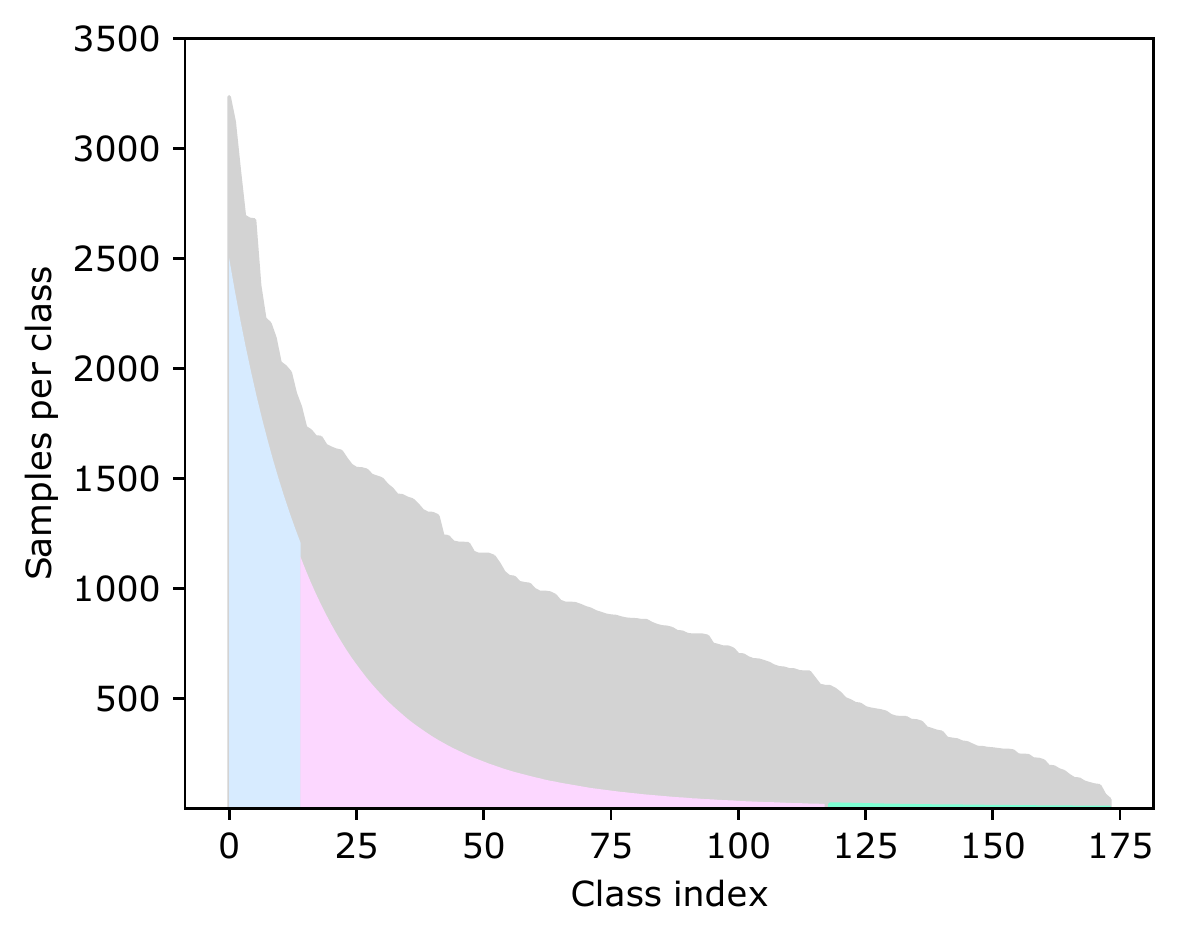}} \subfloat[Something-Something V2 log scale.\label{fig:ssv2constructionlog}]{\includegraphics[width=0.5\linewidth,trim={0 1mm 0 0}]{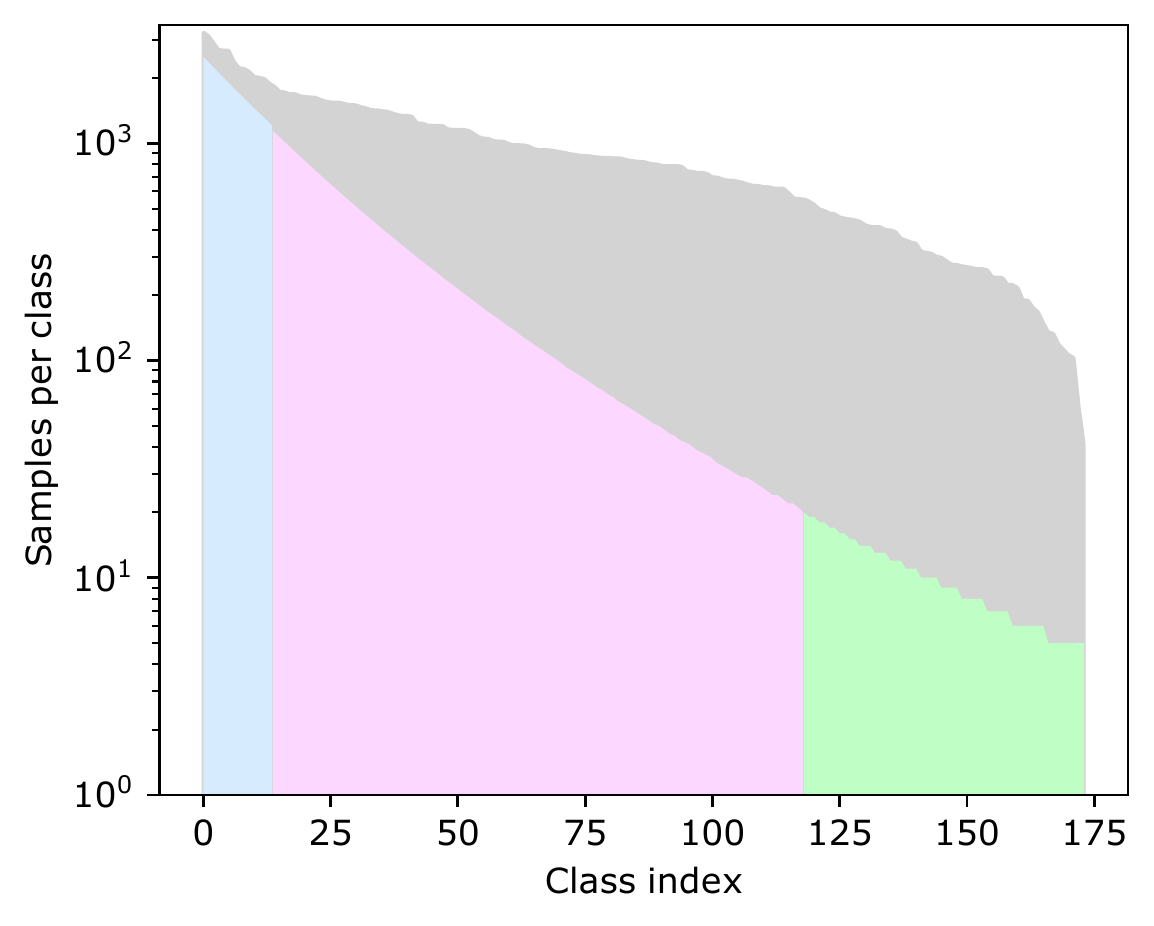}} \\
\subfloat[VideoLT.\label{fig:vltconstruction}]{\includegraphics[width=0.5\linewidth,trim={0 1mm 0 0}]{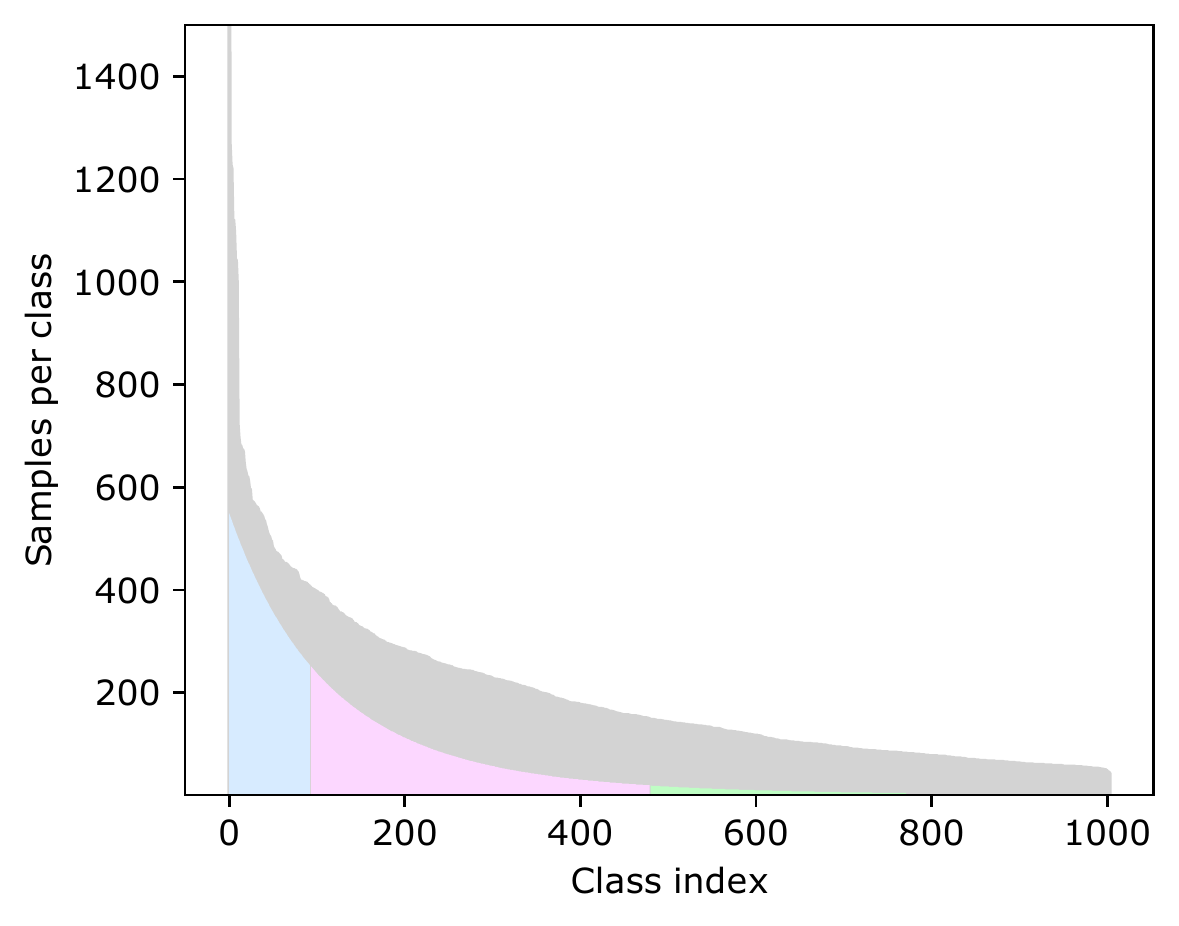}}
\subfloat[VideoLT log scale.\label{fig:vltconstructionlog}]{\includegraphics[width=0.5\linewidth,trim={0 1mm 0 0}]{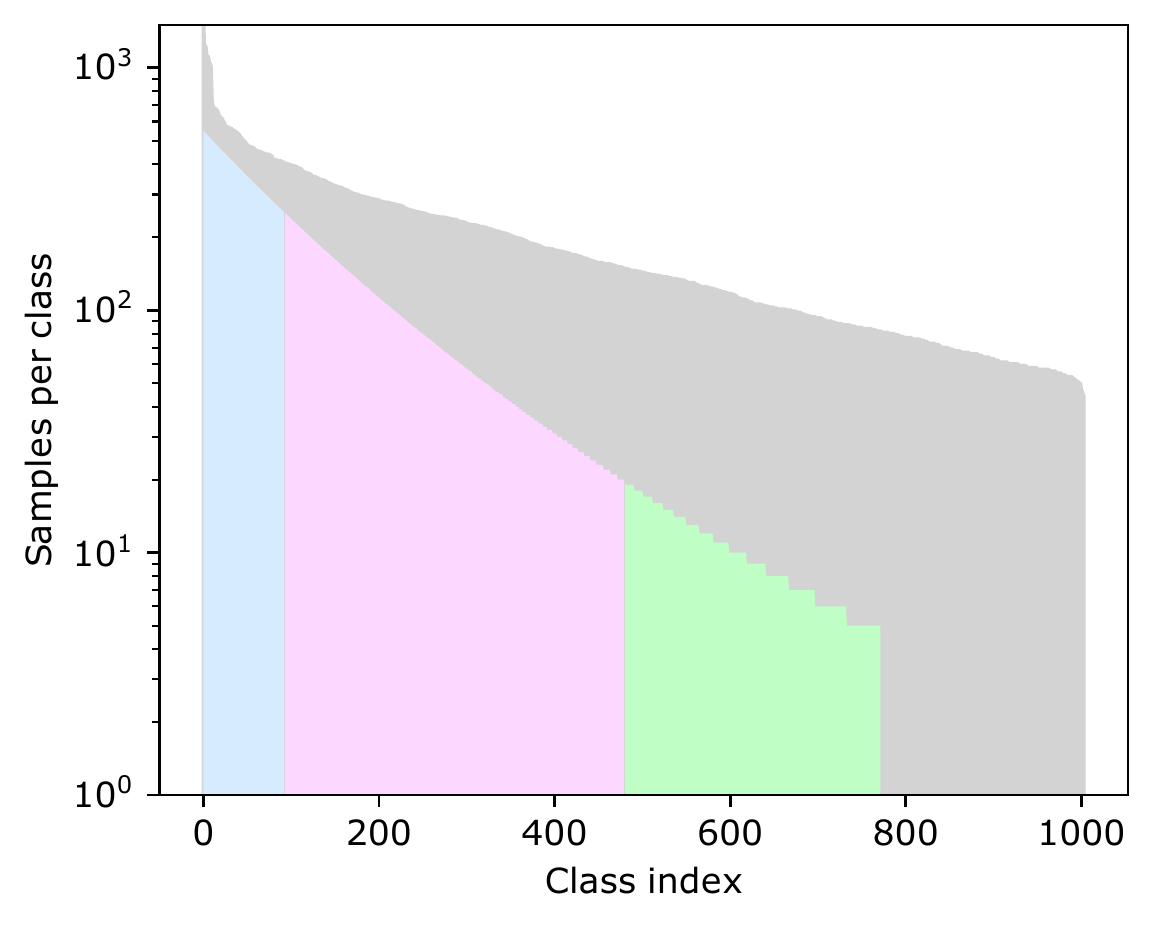}}
\caption{Original datasets (grey) compared to our long-tail versions in standard scale (left) and log scale (right). SSv2 is top and VideoLT is bottom. Blue, pink and green regions show head, tail and few-shot classes in our proposed -LT splits.}
\label{fig:constructions}
\end{figure}

Class count distributions of the original datasets and the (-LT) curated versions are shown in regular and log scale in \cref{fig:constructions}. Splits are shown in Table \ref{tab:constructions}.

\begin{table}[t]
\centering
\footnotesize
\begin{tabular}{@{}lrrrrrrr@{}}
\toprule
           & \multicolumn{3}{l}{ Proposed Properties}           &         &        &       &       \\
           
           \cline{2-4}
{ Dataset}    & {H}\% & {F}\% & {I} & { Cls}. & { Train}  & { Val}   & { Test}  \\ \midrule
SSv2 \cite{Goyal2017}      & 26              & 0               & 79         & 174     & 168913 & 24777 & N/A   \\
SSv2-LT    & 9              & 32              & 500        & 174     & 50418  & 6960  & 2610  \\ \hline
VideoLT \cite{Zhang2021}   & 23              & 0               & 43         & 1004    & 179334 & 25619 & 51239 \\
VideoLT-LT & 12              & 38              & 110        & 772     & 71207  & 7720  & 7720  \\ \bottomrule
\end{tabular}
\vspace{-1mm}
\caption{Original and curated (-LT) long-tail datasets.}
\vspace*{-12pt}
\label{tab:constructions}
\end{table}

\section{Motionformer parameters}\label{app_motionformer}

Table \ref{tab:mf_params} shows the parameter used for Motionformer~\cite{Patrick2021} on EPIC-KITCHENS-100 and SSv2-LT. These are the defaults for EPIC-KITCHENS-100 \cite{Damen2021} and Something-Something V2 \cite{Goyal2017} provided with the code for \cite{Patrick2021}.

\begin{table}[]
\centering
\footnotesize
\begin{tabular}{llll}
\toprule
      & Parameter          & Values   \\ \midrule
Model & Frame size         & {224x224}     \\
      & Num frames         & {16}          \\
      & Num blocks         & {12}          \\
      & Num heads          & {12}          \\
      & Embed dim          & {768}         \\
      & Patch size         & {16}          \\
Train & Input augmentation & {RandAugment} \\
      & Batch size         & {56}          \\
      & Base lr            & {0.0001 instance bal/0.00001 class bal}      \\
      & Momentum           & {0.9}         \\
      & Weight decay       & {0.05}        \\
      & Epochs             & EPIC: 50, SSv2: 35         \\
      & Schedule gamma     & {0.1}         \\
      & Schedule epochs    & EPIC: 30,40, SSv2: 20,30     \\
      & Optimiser          & {adamw}       \\
Test  & Ensemble views     & {10}          \\
      & Spatial crops      & {3}           \\ \bottomrule
\end{tabular}
\caption{Motionformer \cite{Patrick2021} parameters \label{tab:mf_params}}
\end{table}

\end{document}